\documentclass{article}
\usepackage[final]{neurips_2023}

\usepackage[utf8]{inputenc} 
\usepackage[T1]{fontenc}    
\usepackage[pagebackref=true]{hyperref}
\usepackage{url}            
\usepackage{booktabs}       
\usepackage{amsfonts}       
\usepackage{nicefrac}       
\usepackage{microtype}      
\usepackage{xcolor}         
\usepackage{graphicx}
\usepackage{float}
\usepackage{amsmath}

\title{Logit Reweighting for Topic-Focused Summarization}
\author{%
  Joschka Braun\thanks{Equal contribution.} \\
  \And
  Bálint Mucsányi\footnotemark[1] \\
  \And
  Seyed Ali Bahrainian \\
}
\begin{document}
\maketitle

\begin{abstract}
Generating abstractive summaries that adhere to a specific topic remains a significant challenge for language models. While standard approaches, such as fine-tuning, are resource-intensive, simpler methods like prompt engineering often struggle to maintain topical focus, particularly with smaller models. To address this, we propose a lightweight method that enhances topical relevance by directly reweighting the logits of topic-relevant tokens during generation. We evaluate three such reweighting techniques: Constant Shift, which adds a constant value to logits; Factor Scaling, which multiplies them by a factor; and Threshold Selection, which selectively boosts logits that exceed a probability threshold. Experiments on the NEWTS topical summarization dataset, using both Gemma-2B and Llama-3-8B models, show that these techniques effectively increase the use of topic-relevant vocabulary. Notably, the Threshold Selection method successfully improves topical focus without compromising summary quality—a trade-off often seen in other approaches. Our findings demonstrate that directly reweighting logits is a practical and resource-efficient alternative to fine-tuning, offering a promising pathway for precisely controlling the thematic content of generated text.
\end{abstract}

\section{Introduction}
Abstractive topical summarization with autoregressive transformer-based language models (LMs) presents significant challenges, particularly when generating topic-relevant summaries without extensive model retraining~\citep{CATS_Customizable_Abstractive_Topic-based_Summarization, controllable_topic-focussed_abstractive_summarization}. To generate a high-quality summary of a given text and focusing on a selected topic usually requires in-context learning~\citep{In-context_learning}, if the model was not previously fine-tuned to do so. Any model solely fine-tuned for summarization, or a medium-sized instruction-tuned LM, will struggle to generate focused summaries based solely on in-context learning. Fine-tuning models for topical-summarization via methods such as Direct Preference Optimization (DPO)~\citep{Direct_Preference_Optimization} or Reinforcement Learning from Human Feedback (RLHF)~\citep{RLHF} while effective, are technically challenging, require extensive annotated data, and frequently fail to maintain consistent effectiveness across different topics and articles. Therefore, we aim to establish a more easily implementable, universally applicable, and resource-efficient method that improves the topical focus of summaries without extensive training.

\section{Experiment Setup}
\subsection{Datasets and Model}
We evaluate our methods on the NEWTS dataset by \cite{NEWTS_dataset}, designed specifically for topical summarization. The NEWTS dataset is based on the CNN/DailyMail dataset~\citep{CNN_Daily_Mail_dataset} and made up of 2400 training and 600 test samples. Each sample contains a source article and two reference topical summaries, each focused on one of the two most prominent topics in the article (tid1 and tid2). The associated LDA model has 250 distinct topics. For our experiments, we use two state-of-the-art transformer models: the 2 billion-parameter Gemma model by Google~\citep{Gemma_model} and the 8 billion-parameter Llama-3 model from Meta~\citep{Llama3_model}. Both models are pre-trained and have been fine-tuned with instructional data.
\begin{figure}[H]
    \centering
    \begin{minipage}{0.49\textwidth}
        \includegraphics[width=\linewidth]{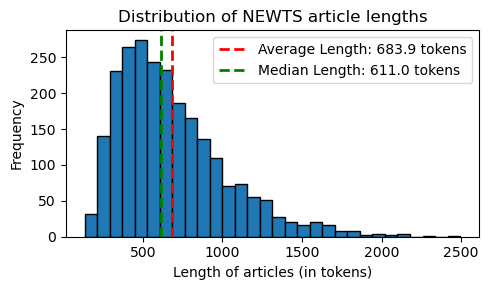}
    \end{minipage}\hfill
    \begin{minipage}{0.49\textwidth}
        \includegraphics[width=\linewidth]{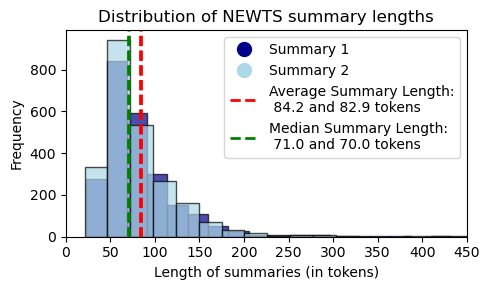}
    \end{minipage}
    \caption{Distribution of NEWTS article and summary lengths. Most articles are between 250 and 1500 tokens long. The summaries are roughly 12\% as long as the original articles.}
    \label{fig:newts_dataset_article_and_summary_lengths}
\end{figure}
\subsection{Hyperparameters}
For all experiments conducted, we maintained the same hyperparameters to ensure comparability across different methods and models. We used the NEWTS training dataset, selecting articles in chronological order for consistency. The number of articles varied between 25 or 50 based on the size of the model being tested or the use of beam search to accommodate computational constraints and model capabilities. The length of the generated summaries was limited between 80 and 90 tokens, which is roughly similar to the mean summary length of the human-written summaries of the NEWTS dataset. This length was chosen to make summaries comparable to the human-written summaries and balance detail and brevity. For context, the average article length in the dataset is approximately 680 tokens, so summaries are around 12\% in length or the original article.

Nucleus sampling was implemented with a top-p parameter of 0.95 to restrict the model’s selection to the most likely subsequent tokens. Additionally, the sampling was restricted by a top-k value of 50. In experiments without beam search, generation was solely based on this sampling strategy. For the experiment involving beam search, we utilized a configuration of 4 beams to diversify the exploration of potential summary outputs.

From the LDA model, we consistently used the top 25 words associated with the topic to represent it. To translate from words to tokens, we generated multiple variations of each word, including the lemmatized and stemmed versions, as well as variations in capitalization and spacing. Instead of 1–2 tokens associated with a word, we had 3–5. This step is crucial for covering all possible tokens that should be more likely to be sampled when the model is generating a summary with enhanced topical focus on that specific word. 

Overall, our experiment hyperparameters were chosen to generate high-quality summaries and make generated summaries comparable to the human-written summaries in the NEWTS dataset. Keeping hyperparameters identical ensures that any observed differences in performance are attributable to the methodological variations rather than differences in experimental conditions. 

\subsection{Evaluation methods}
To evaluate the generated summaries, we evaluate for both the quality of the summary, compared to the respective reference summary from the NEWTS dataset. And we evaluate for topical focus respective to the chosen topic with respect to the topic of the LDA model. To rigorously assess the performance of the generated summaries, we employ three metrics for each the quality and topical focus of the summary. This dual assessment approach ensures a comprehensive evaluation of how well the summaries adhere to both the linguistic quality and relevance to the specified topic.

\newpage
\subsubsection{Summarization Quality}
To quantify summarization quality, we utilize the following metrics:

\textbf{ROUGE-L Score:} ROUGE-L (Recall-Oriented Understudy for Gisting Evaluation - Longest Common Subsequence) introduces by \cite{ROUGE_score} measures the longest common subsequence between the generated summary and a reference summary. It is particularly useful for evaluating the summary for fluency and coherence. We use the implementation from the \textbf{`rouge\textunderscore scorer.RougeScorer`} class and active stemming.\\
\textbf{MAUVE Score:} MAUVE (Measure of Aggregated Unidirectional Validity and Entropy) introduced by \cite{MAUVE_score} quantifies the statistical gap between the human-written and machine-generated summaries using the Kullback-Leibler divergence between the two distributions in the embedding space. The MAUVE score correlates well with human evaluations for text quality. We use the official implementation of MAUVE via a wrapper from the \textbf{`evaluate`} library on Hugging Face.\\
\textbf{BERTScore:} BERTScore~\citep{BERT_score} leverages the pre-trained contextual embeddings, originally from BERT~\citep{BERT_model} models, to compute the semantic similarity between the generated summary and the reference text. This metric is robust against paraphrasing and variations in word choice, thus offering a nuanced measure of semantic similarity. We employ the \textbf{`BERTScorer`} class with the microsoft/deberta-xlarge-mnli model~\citep{Deberta_model}, as it currently correlates best with human evaluation.\\
Collectively, these three metrics offer a robust estimate of the summarization quality.
\vspace{-0.1cm}
\subsubsection{Summary Topical Focus}
To evaluate the alignment of generated summaries with the intended topics, we utilize three methods:
\textbf{Lemmatization-Based Evaluation:} This method processes the generated text by lemmatizing words, converting them to their canonical form. Using the LDA model, it matches these lemmas against the lemmas of the top topic words identified for the relevant topic. The topical focus score is then calculated as the weighted presence of these lemmas in the summary, normalized by the total weight of topic-specific lemmas.\\
\textbf{Tokenizing-Based Evaluation:} In this approach, text is tokenized using a model-specific tokenizer, and tokens are matched against those generated from topic-specific words identified by the LDA model. The score is computed based on the proportion of topic tokens in the summary, offering a direct measure of topical vocabulary usage.\\
\textbf{Dictionary-Based Evaluation:} This method leverages a dictionary approach where each word in the summary is converted into its bag-of-words representation. The LDA model then provides a distribution over topics for these words. The score reflects the prevalence of the relevant topic's tokens within the summary, adjusted by the topic distribution provided by the LDA.

By integrating these three measures, we can estimate the effectiveness of the logits reweighting methodologies in producing topically focused summaries.

\section{Abstractive Topical Summarization via Prompt Engineering}
\subsection{Method}
To establish a baseline, we employ prompt engineering to explicitly direct the model's attention to specific topics. The prompt begins with the instruction to generate a summary, which sets the task context for the language model. To direct the model's focus towards the desired topic, the prompt incorporates a specific instruction to concentrate on the topic, delineated by the top 25 words associated with it. These topic-associated words are integrated into the prompt to ensure that the model focuses on the specific topic during the summary generation.

\subsection{Results}
\begin{figure}[H]
    \centering
    \begin{minipage}{0.49\textwidth}
        \includegraphics[width=\linewidth]{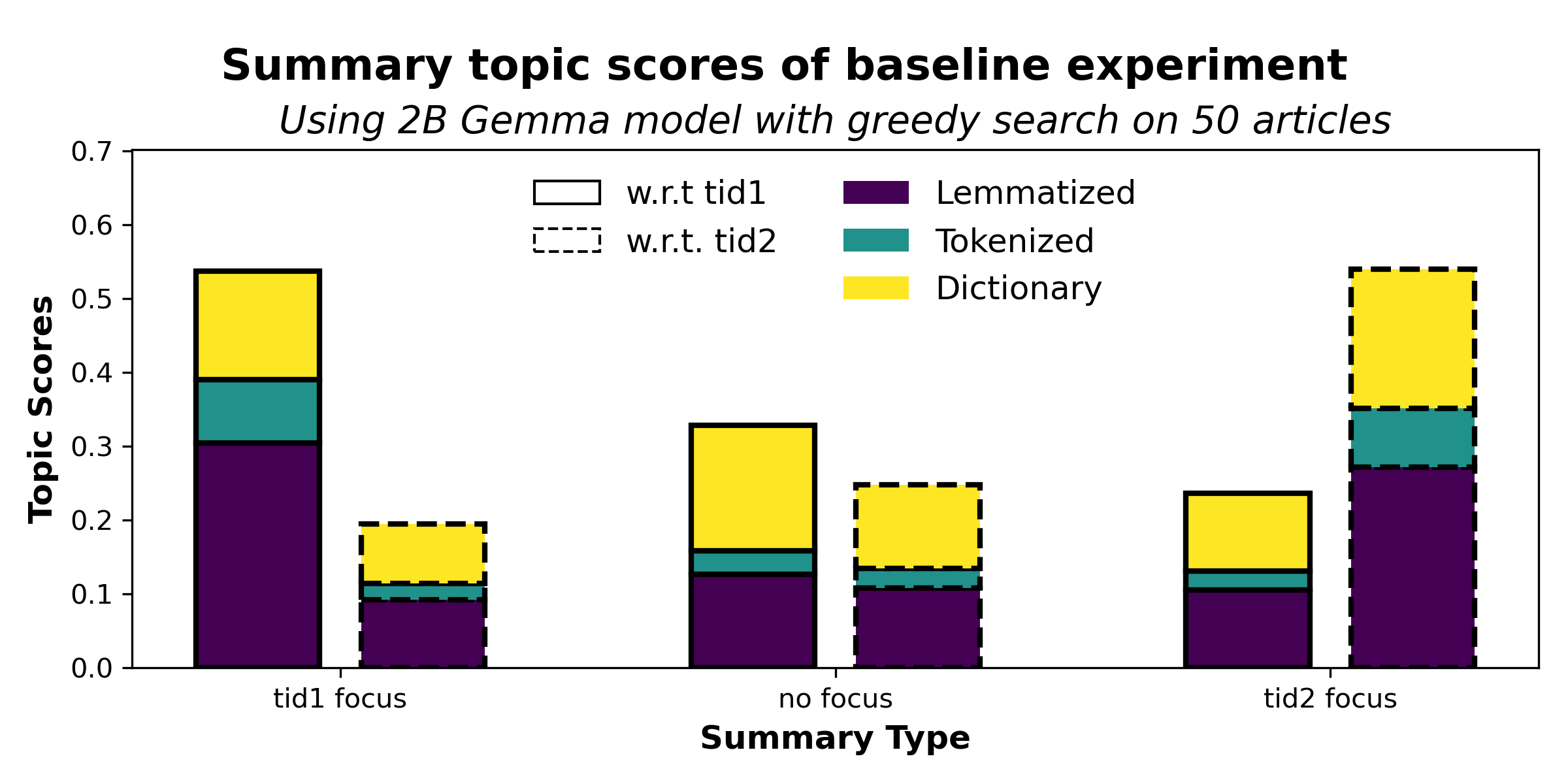}
    \end{minipage}\hfill
    \begin{minipage}{0.49\textwidth}
        \includegraphics[width=\linewidth]{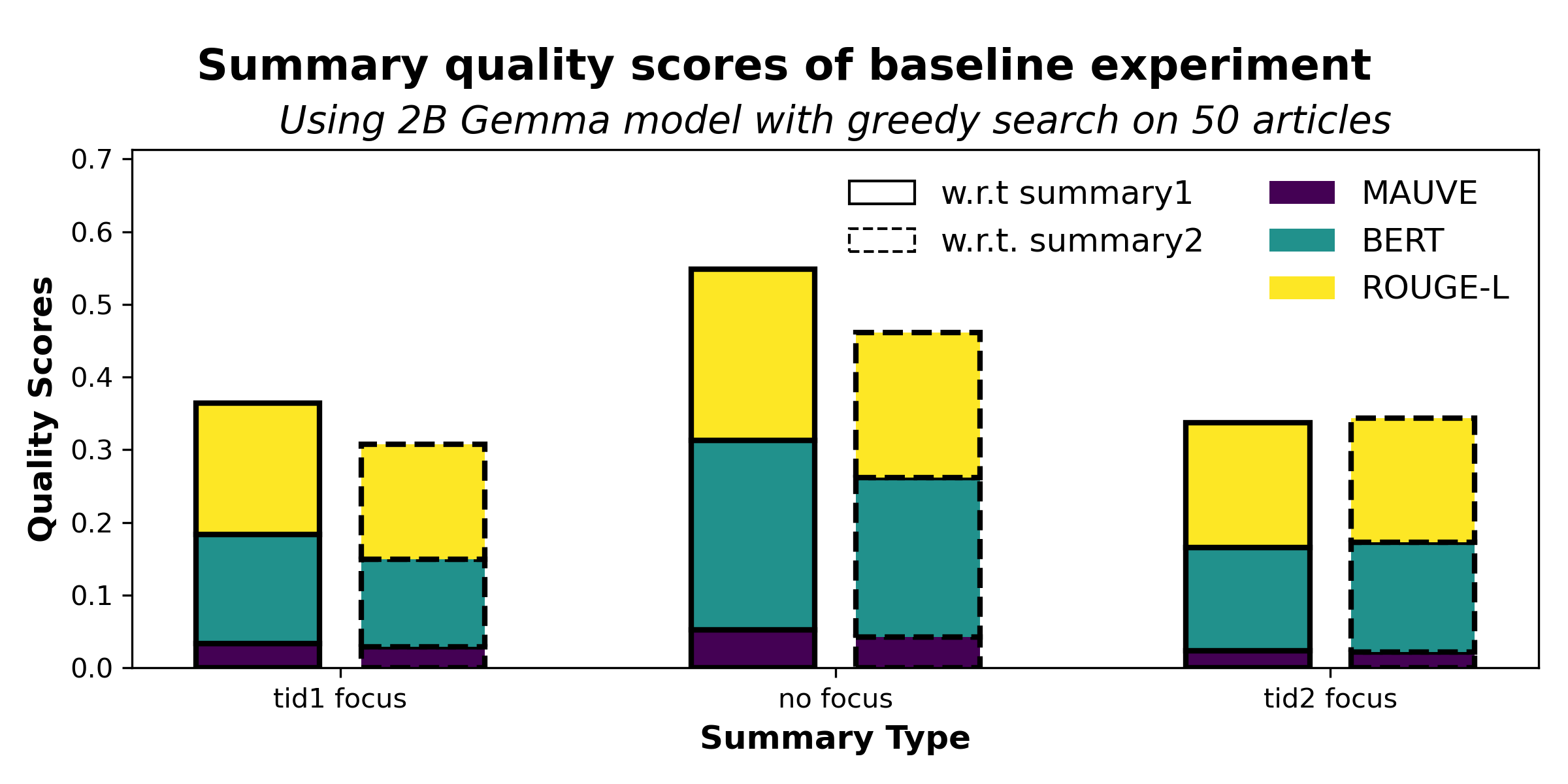}
    \end{minipage}
    \caption{Summary scores for the baseline experiment using Gemma-2B with greedy decoding. Prompt engineering effectively changes topical focus, but has a measurable negative side effect on summary quality.}
    \label{fig:baseline_gemma_2b_greedy}
    \vspace{-0.3cm}
\end{figure}
\begin{figure}[H]
    \centering
    \begin{minipage}{0.49\textwidth}
        \includegraphics[width=\linewidth]{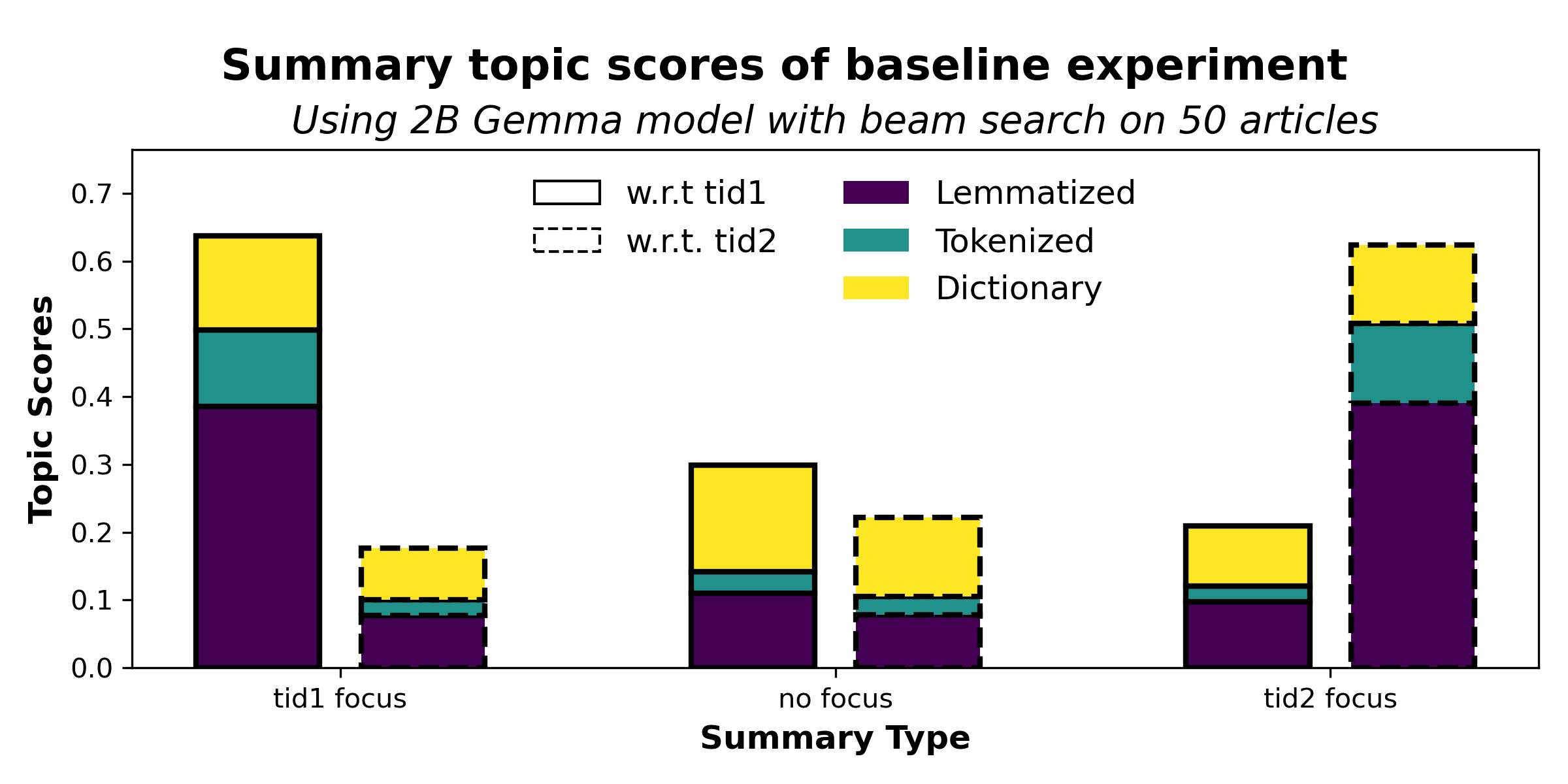}
    \end{minipage}\hfill
    \begin{minipage}{0.49\textwidth}
        \includegraphics[width=\linewidth]{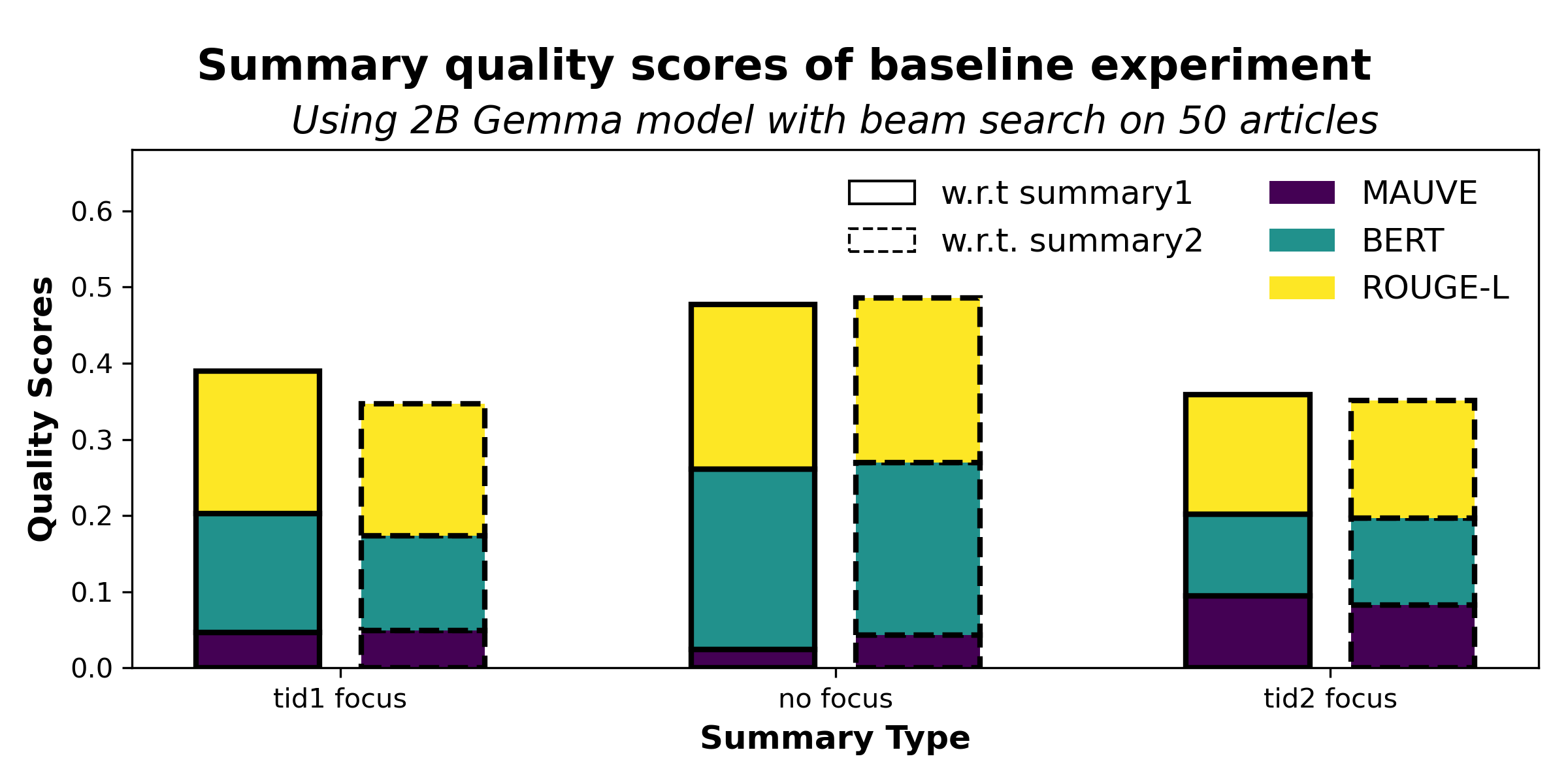}
    \end{minipage}
    \caption{Summary scores for the baseline experiment using Gemma-2B with beam search. Decoding with beam search both improves the resulting topical focus and mitigates degradation in summary quality compared to greedy decoding.}
    \label{fig:baseline_gemma_2b_beamsearch}
    \vspace{-0.3cm}
\end{figure}
\begin{figure}[H]
    \centering
    \begin{minipage}{0.49\textwidth}
        \includegraphics[width=\linewidth]{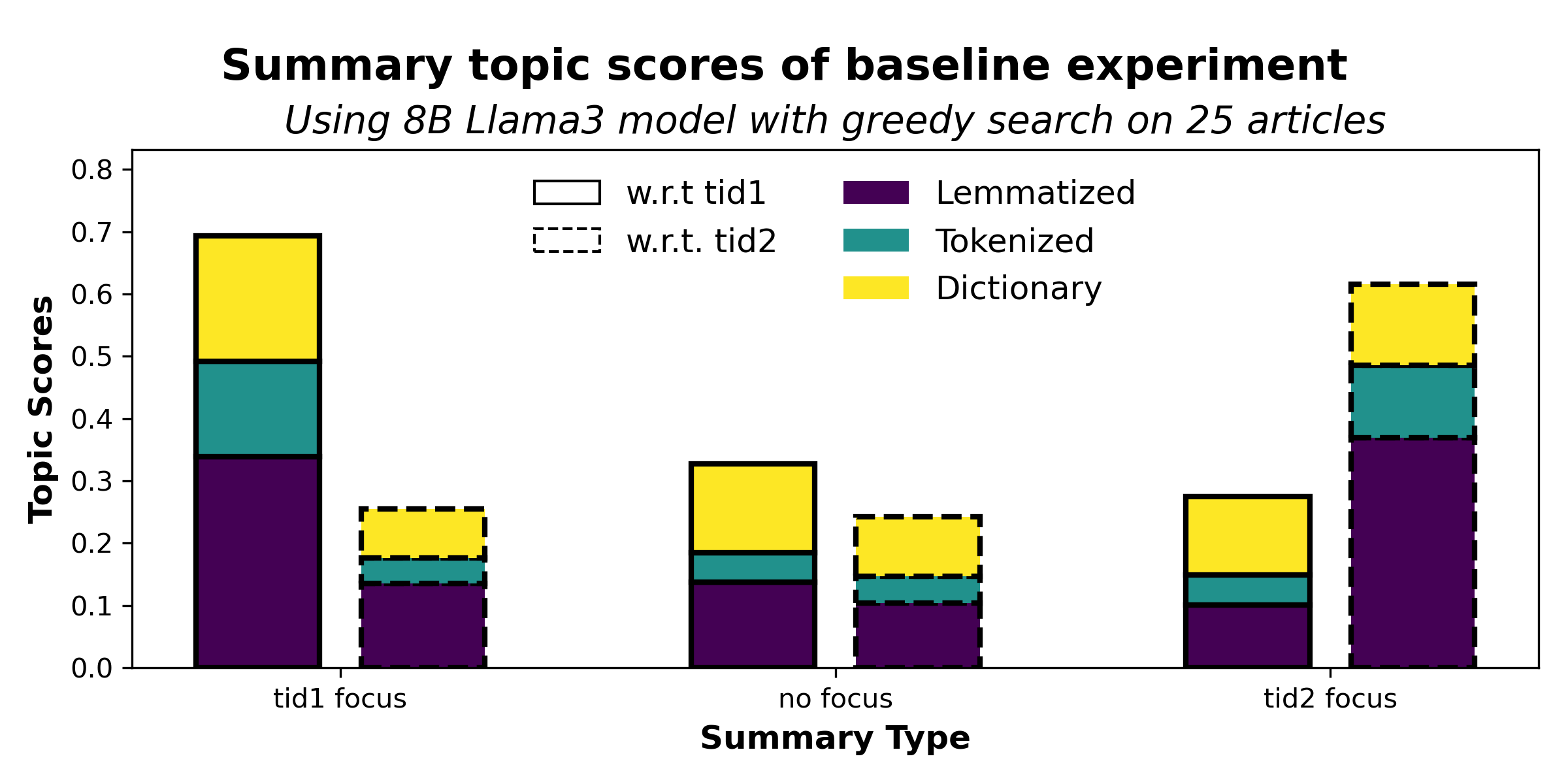}
    \end{minipage}\hfill
    \begin{minipage}{0.49\textwidth}
        \includegraphics[width=\linewidth]{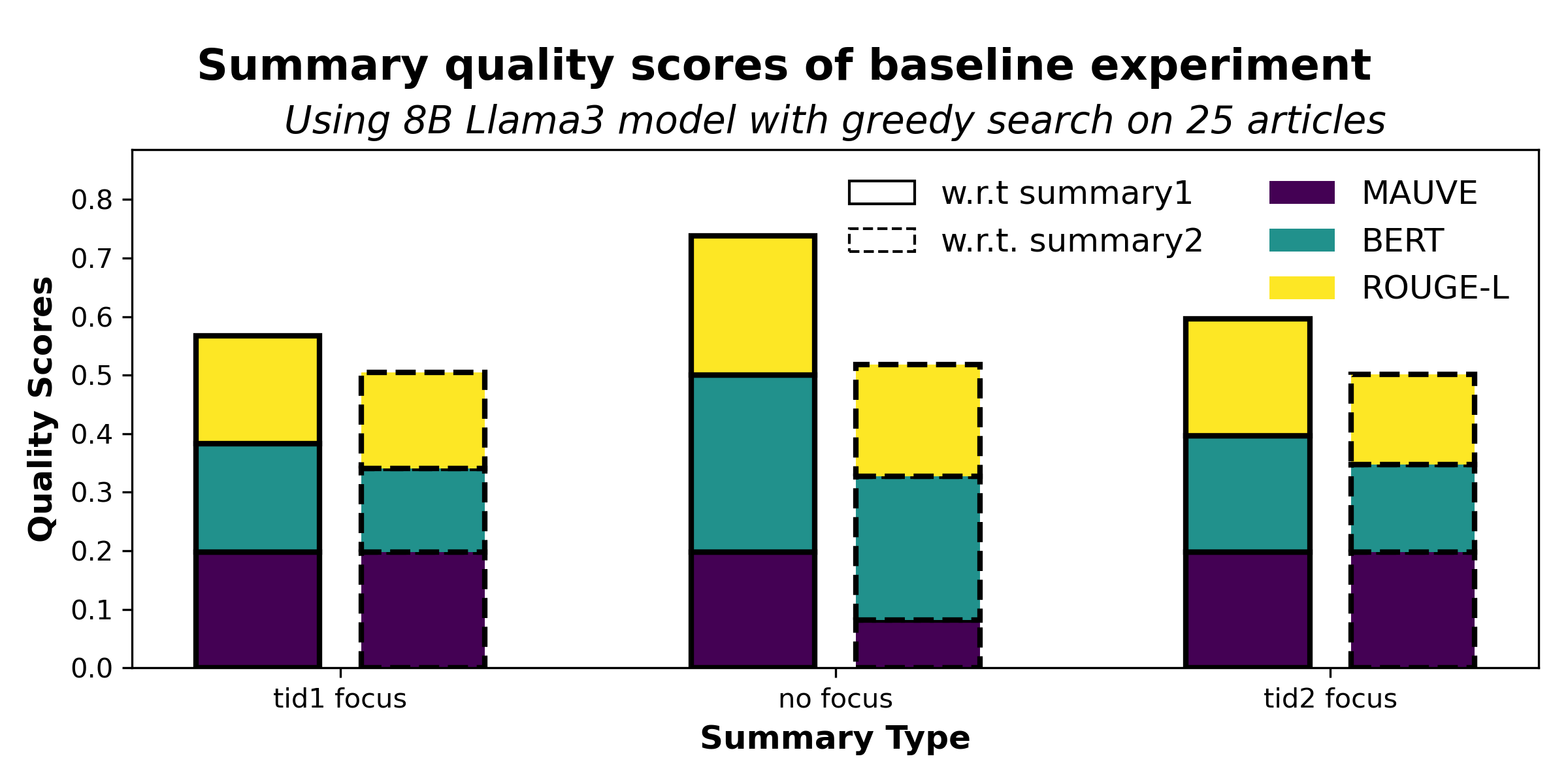}
    \end{minipage}
    \caption{Summary scores for the baseline experiment using Llama-3-8B with greedy decoding. The larger Llama model has the highest topical focus and summary quality.}
    \label{fig:baseline_llama_8b}
\end{figure}
We observed consistent trends across both the Llama-3-8B and Gemma 2B models with and without beam search. When prompted to focus on a specific topic, the topical scores increased roughly twofold in all three setups, particularly the lemmatization and tokenization-based topic scores. When prompting the model to focus on the respective topic 1, the topic score approximately doubled for topic 1 and decreased noticeably for topic 2 compared to the baseline. The same is also true for summaries focused on topic 2. Changes in topical scores are larger when using beam search compared to greedy decoding, or using the larger Llama-3-8B model compared to the smaller Gemma-2B model. In terms of summary quality, prompting the model to focus on a particular topic came at the expense of overall summary quality. Summaries generated without a specific topical focus consistently measured higher scores. When expecting the generated summaries, this was likely caused due to the model reiterating the instructions at the start of the generated summary. This included reiterating words related to the designated topic, a characteristic behavior of instruction-tuned models, which could not be effectively mitigated by further prompt engineering. In terms of summary quality, both beam search and a larger model size increased absolute summary quality. 
These findings underscore the challenge of balancing topical focus with summary quality in instruction-tuned models and highlight the limitations of prompt engineering in effectively addressing this issue.

\section{Abstractive Topical Summarization via Logits Reweighting}
\subsection{Methods}
To manipulate the logits predicted by the model, we have implemented a custom LogitsProcessor class within the Hugging Face transformer framework. This class enables the alteration of the logits of topic-specific tokens during text generation. Each of our three methods modifies the logits in a distinct manner, allowing for experimental comparisons of their effects on the generated summaries.
\subsubsection{Constant Shift}
The Constant Shift method adds a constant value $c$ to all logits of predefined topic-relevant tokens. This shift uniformly increases or decreases the probability of selecting such tokens, irrespective of their original logits. The modification is implemented as:
$$\text{scores}_{\text{modified}}[i] = \text{scores}[i] + c$$
where $i$ indexes into topic-relevant tokens. This method ensures consistent manipulation of specified tokens across different contexts.

\subsubsection{Factor Scaling}
Factor Scaling alters the logits of topic-specific tokens by multiplying them by a predetermined scaling factor $\alpha$. The method can be expressed mathematically as:
$$\text{scores}_{\text{modified}}[i] = \text{scores}[i] \times \alpha$$
where $i$ denotes topic-relevant tokens. This scaling affects logits proportionally, magnifying or diminishing their original values and thus altering their likelihood of selection. 
\subsubsection{Threshold Selection}
Threshold Selection involves selectively adjusting logits based on a predefined probability threshold $\theta$. Logits are first converted to probabilities using the softmax function. If a topic-relevant token's probability exceeds $\theta$, its logit is increased to the maximum logit in the current logit distribution plus an additional encouragement factor $\beta$. This is expressed as:
$$\text{scores}_{\text{modified}}[i] = 
\begin{cases} 
\max(\text{scores}) + \beta & \text{if softmax}[i] \geq \theta \\
\text{scores}[i] & \text{otherwise}
\end{cases}$$
where softmax[i] is the calculated probability of topic-relevant token i. Threshold Selection only changes token logits that are already likely under the model's current context, enhancing their prominence without affecting less relevant tokens. These methods provide distinct strategies for changing topic relevance during the summary generation process.

\subsection{Results}
\subsubsection{Constant Shift}
\begin{figure}[H]
    \centering
    \begin{minipage}{0.49\textwidth}
        \includegraphics[width=\linewidth]{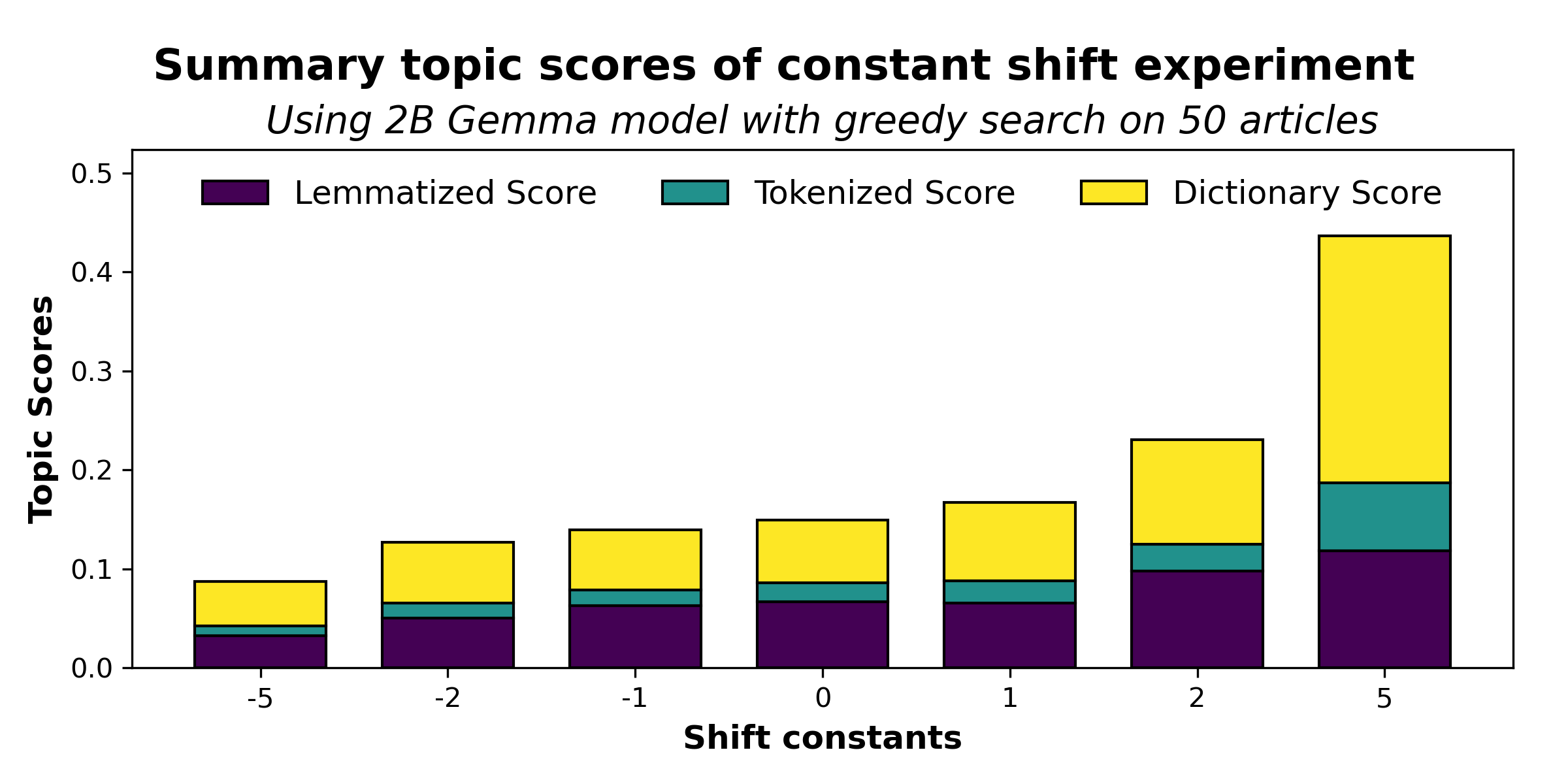}
    \end{minipage}\hfill
    \begin{minipage}{0.49\textwidth}
        \includegraphics[width=\linewidth]{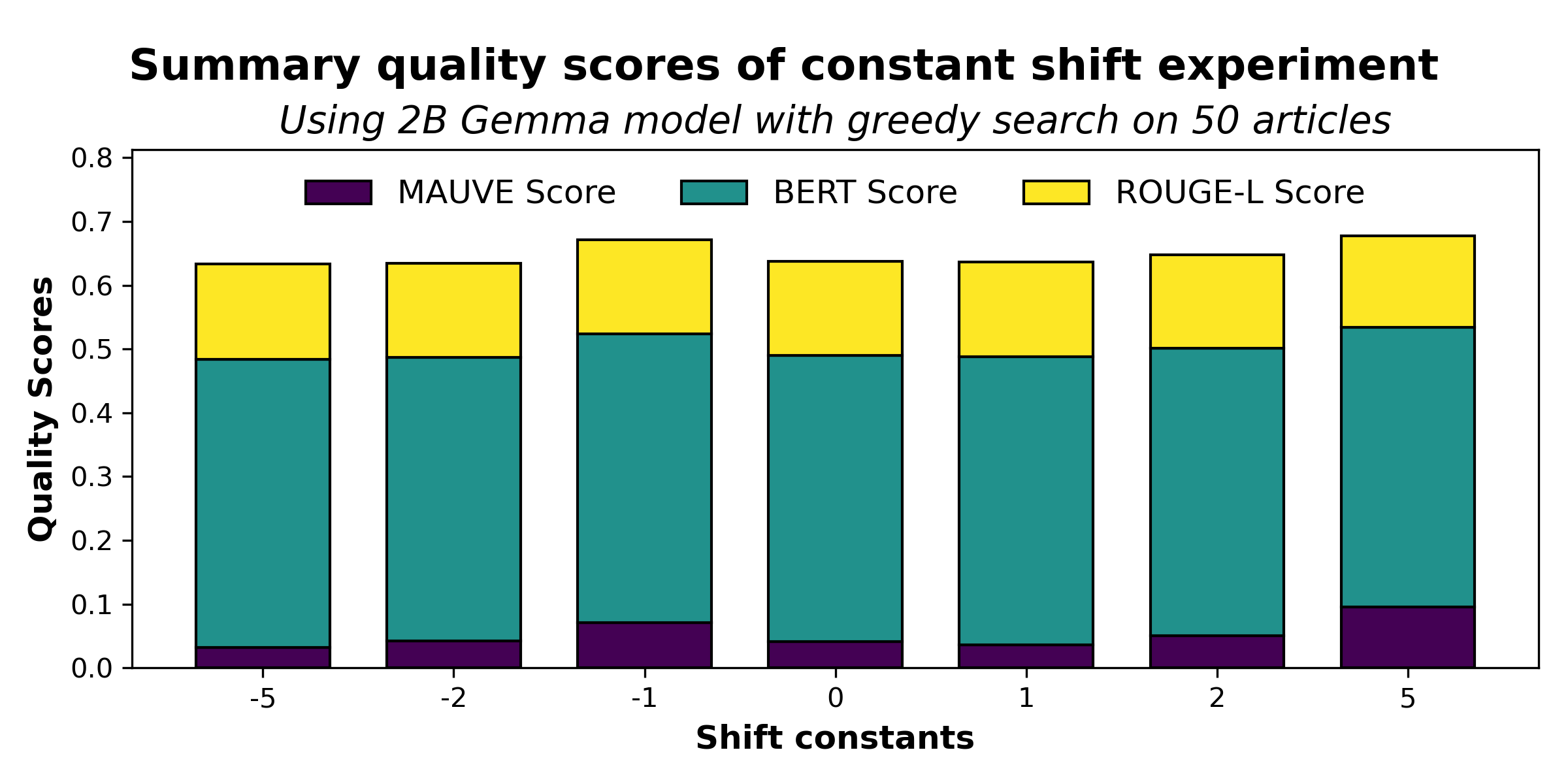}
    \end{minipage}
    \caption{Summary scores for the constant shift experiment using Gemma-2B with greedy decoding. Adding or subtracting a constant shift to topic-relevant tokens effectively changes the topical scores without negatively impacting quality scores.}
    \label{fig:constant_shift_gemma_2b_greedy}
    \vspace{-0.3cm}
\end{figure}
\begin{figure}[H]
    \centering
    \begin{minipage}{0.49\textwidth}
        \includegraphics[width=\linewidth]{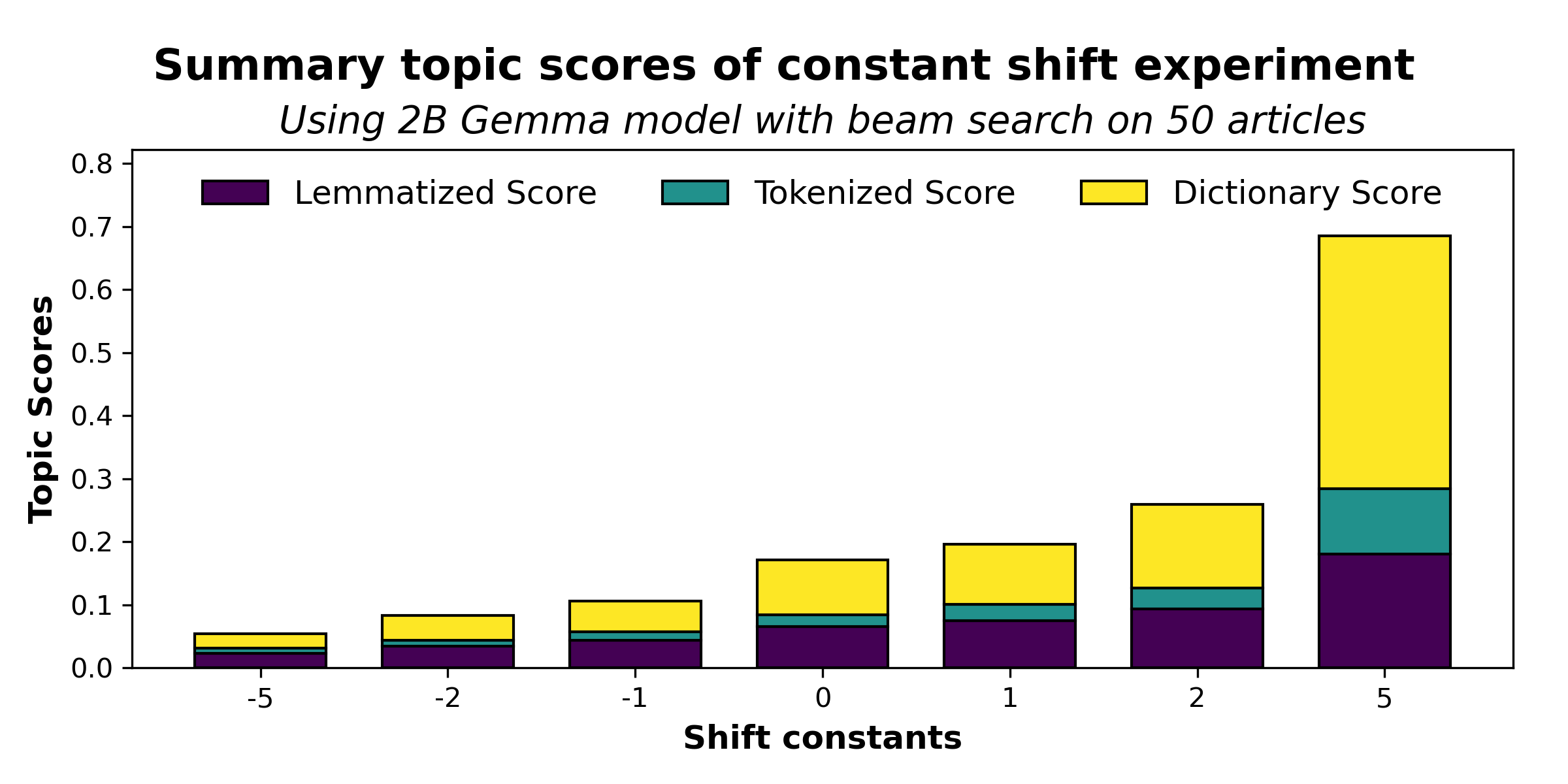}
    \end{minipage}\hfill
    \begin{minipage}{0.49\textwidth}
        \includegraphics[width=\linewidth]{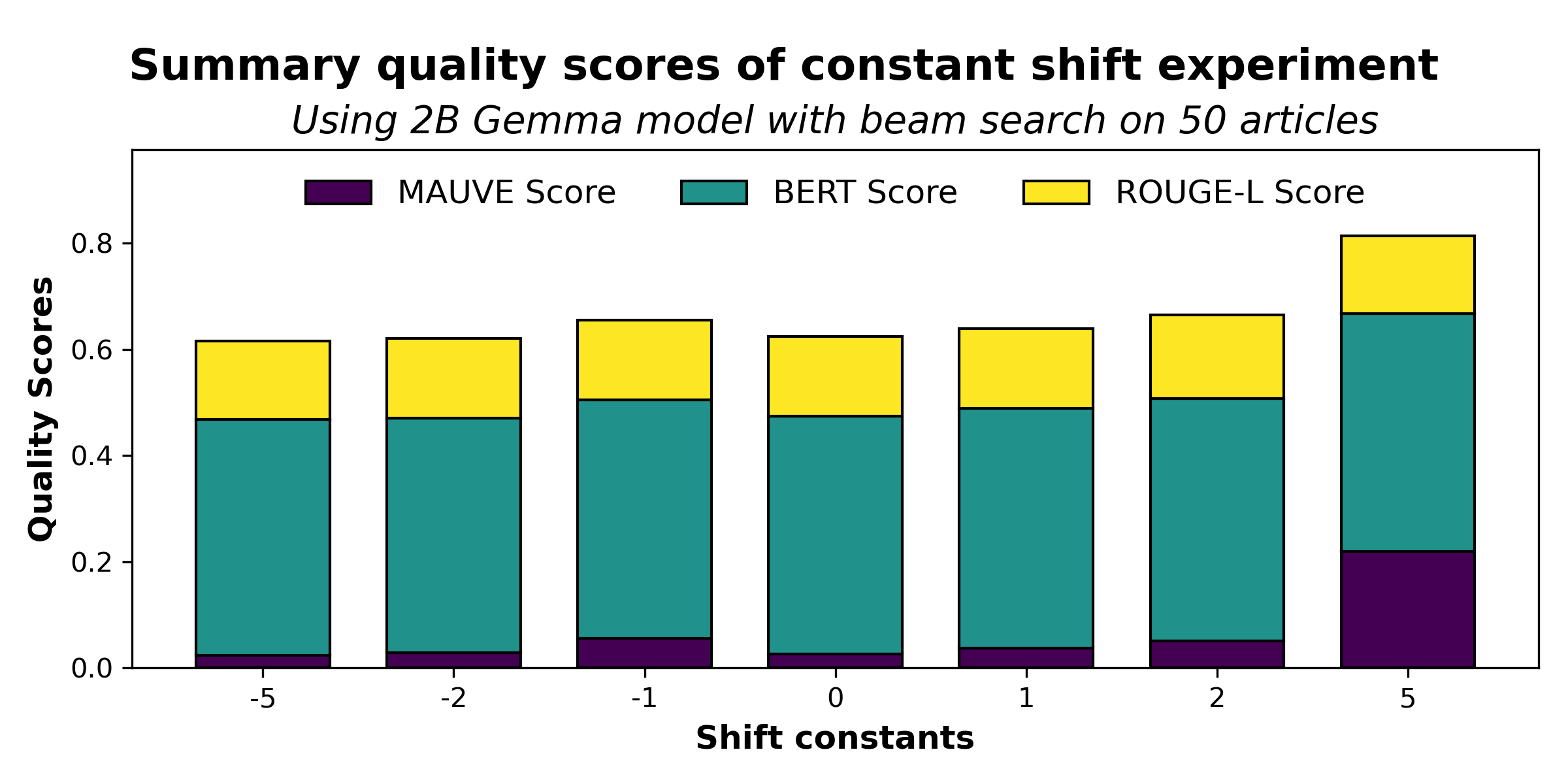}
    \end{minipage}
    \caption{Summary scores for the constant shift experiment using Gemma-2B with beam search. Beam search has a similar pattern to the greedy search results for the Gemma-2B model. Adding or subtracting a constant value even increases quality scores relative to reference summaries.}
    \label{fig:constant_shift_gemma_2b_beamsearch}
    \vspace{-0.3cm}
\end{figure}
\begin{figure}[H]
    \centering
    \begin{minipage}{0.49\textwidth}
        \includegraphics[width=\linewidth]{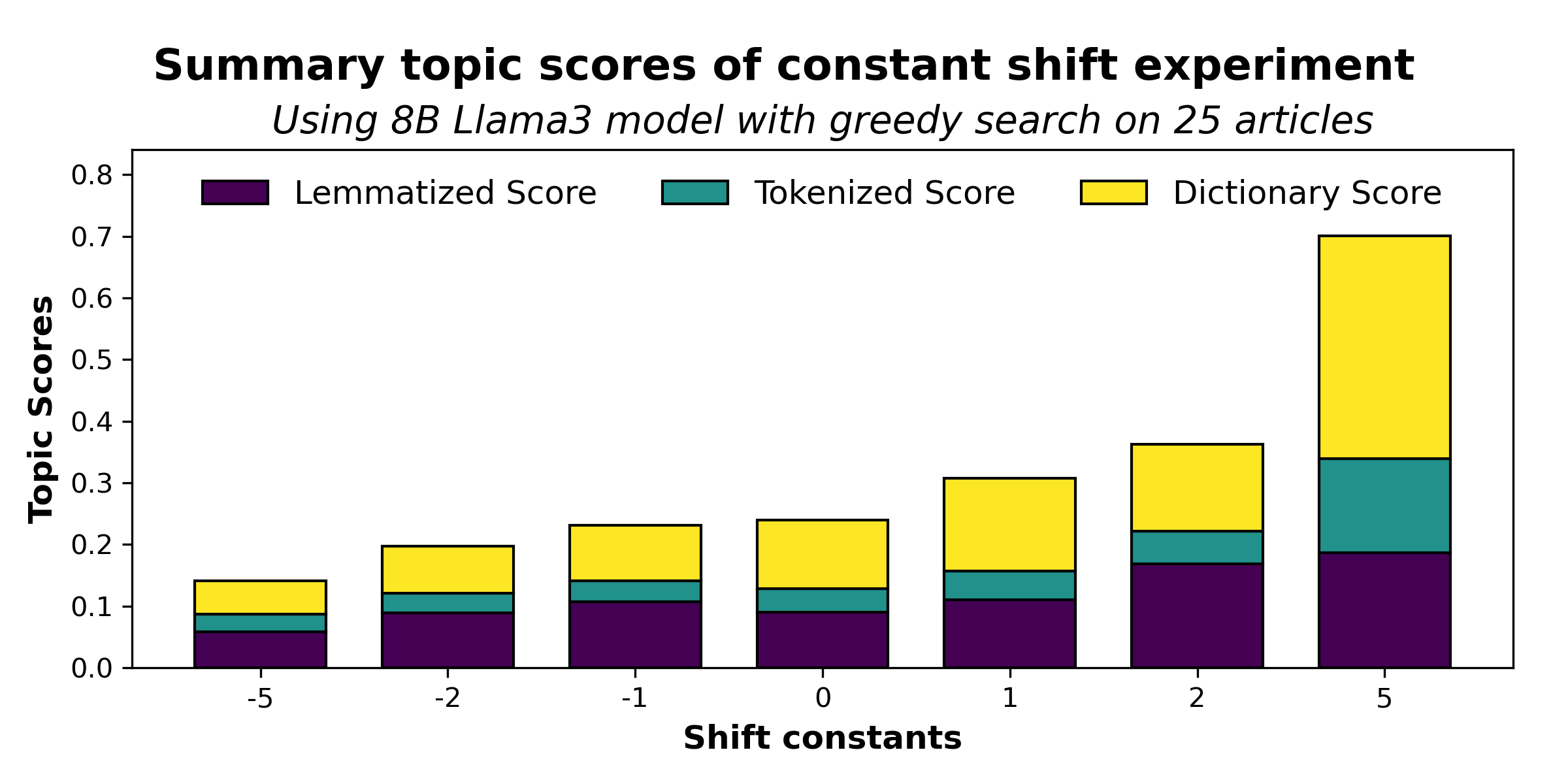}
    \end{minipage}\hfill
    \begin{minipage}{0.49\textwidth}
        \includegraphics[width=\linewidth]{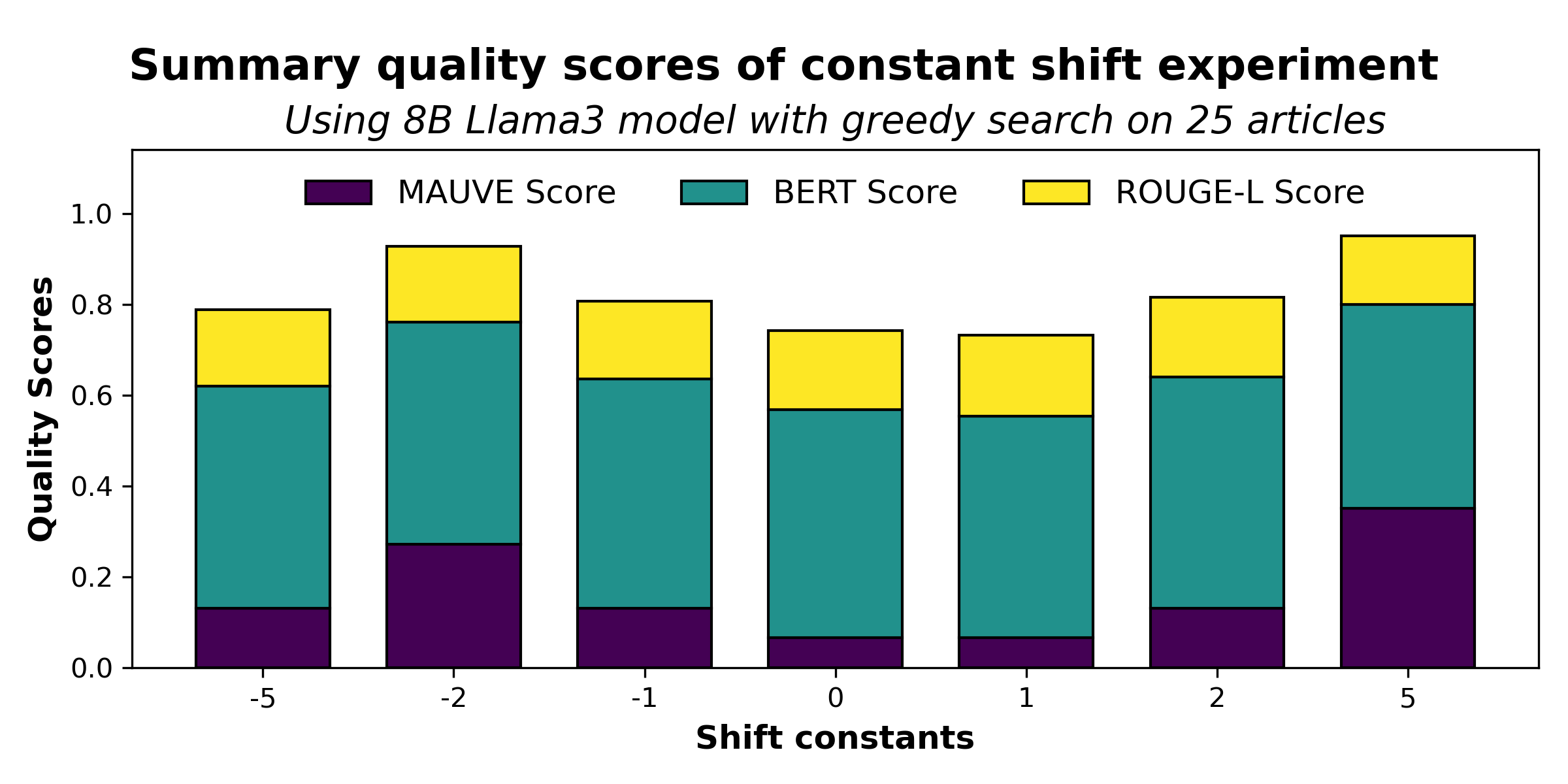}
    \end{minipage}
    \caption{Summary scores for the constant shift experiment using Llama-3-8B with greedy decoding. As for prompting, the Llama model has the highest topic and quality scores overall.}
    \label{fig:constant_shift_llama_8b}
    \vspace{-0.3cm}
\end{figure}
In the Constant Shift experiment, a consistent pattern emerged across both the Llama and Gemma models with greedy decoding and beam search. Adjusting topic-relevant token logits by adding constant values had a significant impact on topical scores, as measured by lemmatization, tokenization, and dictionary-based evaluations. Specifically, negative constants decreased topical scores, while positive values increased them across all metrics. This effect was more pronounced with beam search and in the larger Llama-3-8B model compared to the Gemma-2B model. ROUGE-L and BERT scores remained stable despite these modifications; however, MAUVE scores showed considerable variability, peaking notably at a shift constant of 5. This variance underscores the sensitivity of the MAUVE metric to changes in topical emphasis. Despite this, the Llama model consistently demonstrated superior summary quality. As tested in other experiments, if the shift constant exceeds 10, summaries are composed almost exclusively of topic-relevant tokens, severely compromising summary quality in terms of fluency and coherence. This underscores that while topical encouragement via constant shift is successful, the shift constant should not be chosen too large.
\subsubsection{Factor Scaling}
\begin{figure}[H]
    \centering
    \begin{minipage}{0.49\textwidth}
        \includegraphics[width=\linewidth]{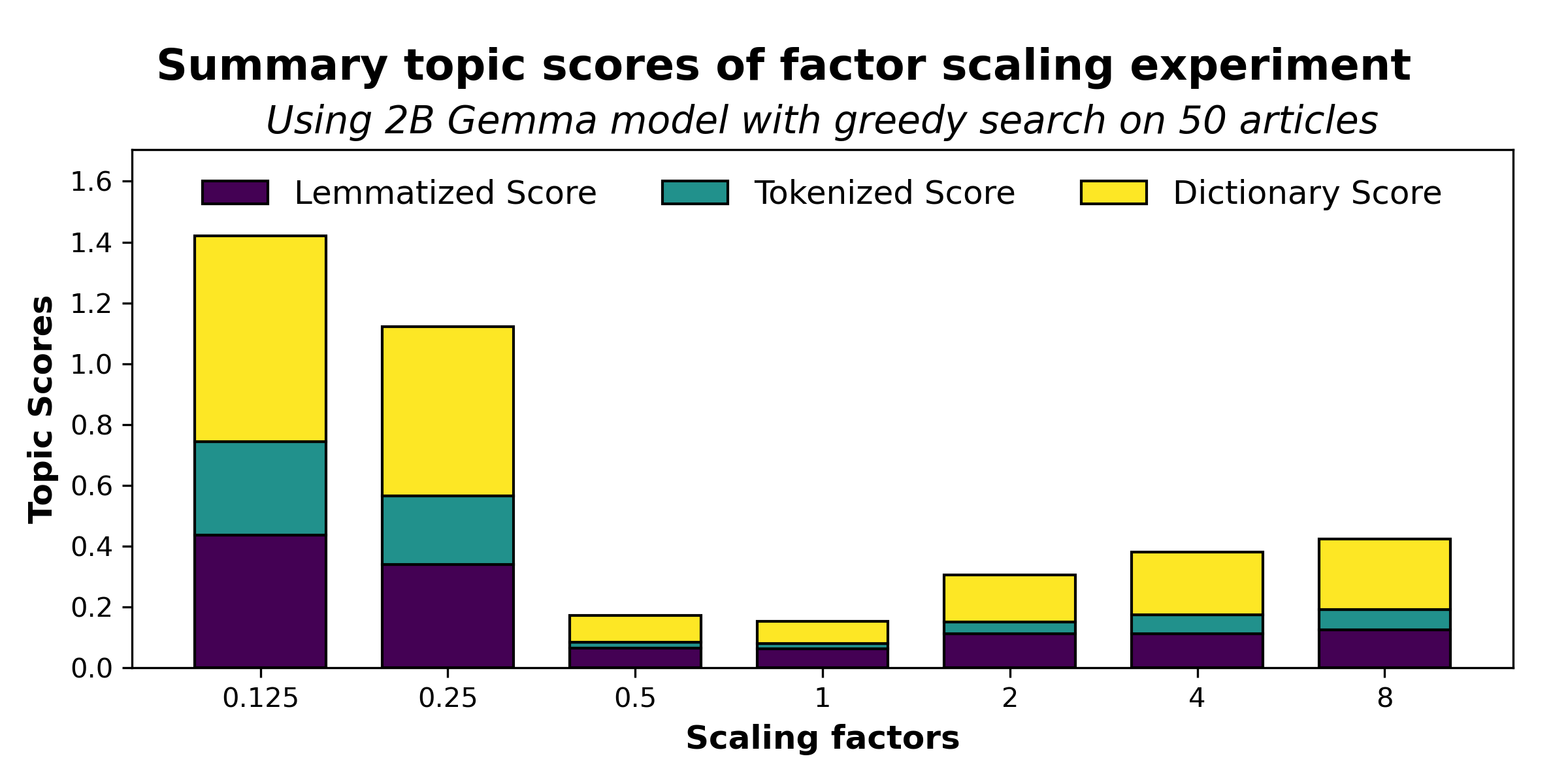}
    \end{minipage}\hfill
    \begin{minipage}{0.49\textwidth}
        \includegraphics[width=\linewidth]{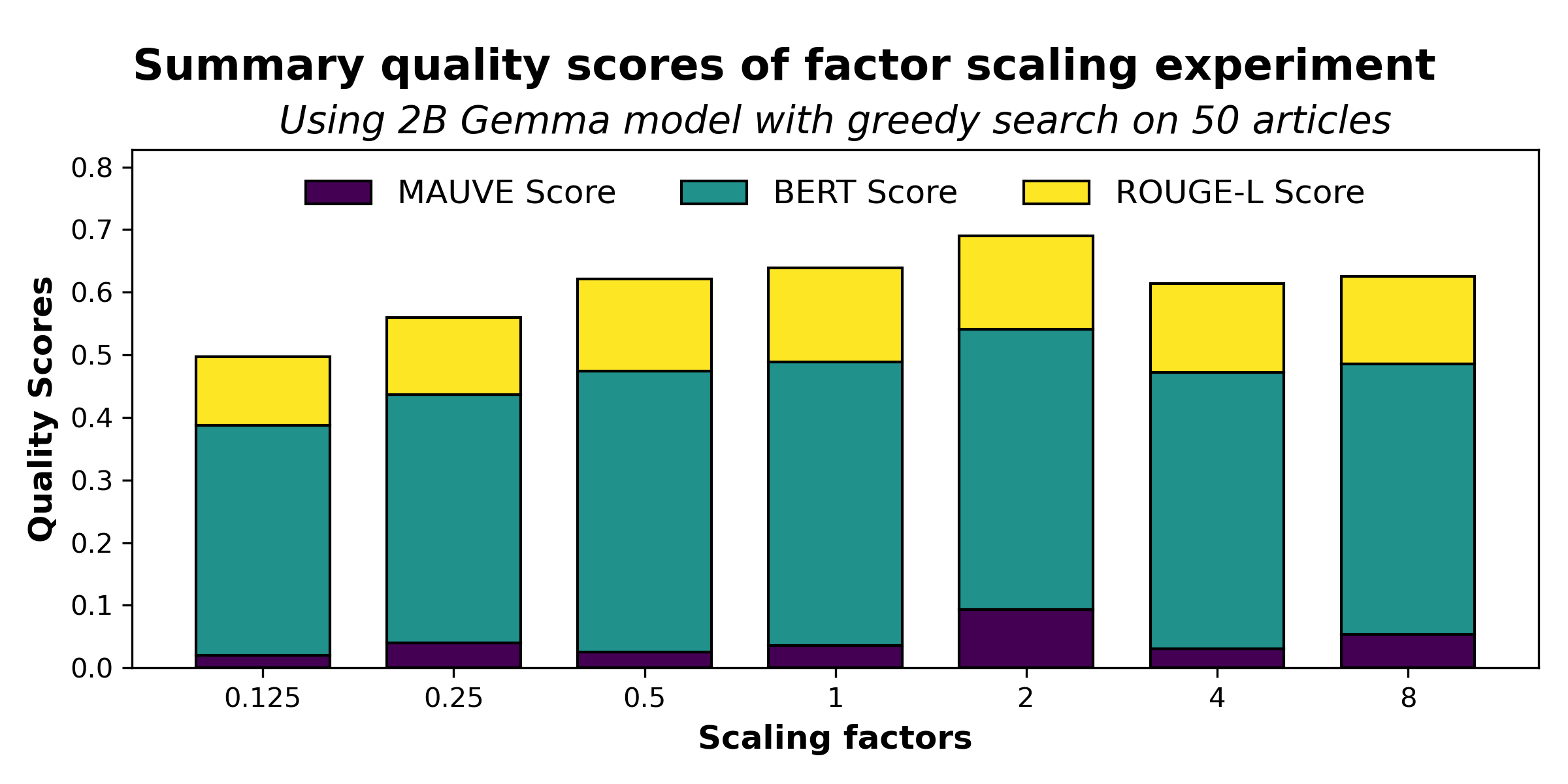}
    \end{minipage}
    \caption{Summary scores for the factor scaling experiment using Gemma-2B with greedy decoding. The overall impact on topical focus is largest for very small factors, exceeding the impact that the constant shift has. However, this increase in topical focus also leads to lower summary quality scores.}
    \label{fig:factor_scaling_gemma_2b_greedy}
    \vspace{-0.3cm}
\end{figure}
\begin{figure}[H]
    \centering
    \begin{minipage}{0.49\textwidth}
        \includegraphics[width=\linewidth]{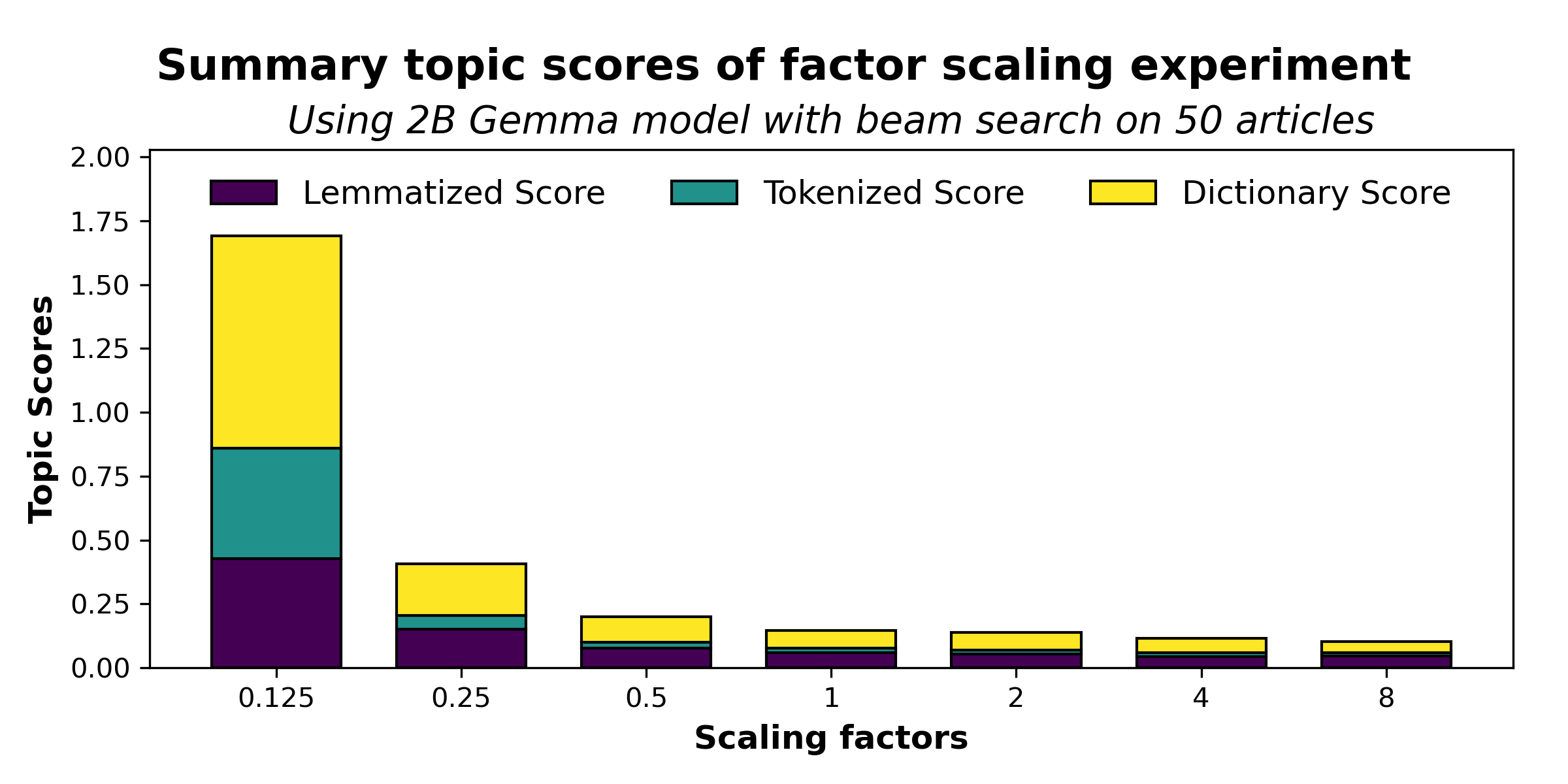}
    \end{minipage}\hfill
    \begin{minipage}{0.49\textwidth}
        \includegraphics[width=\linewidth]{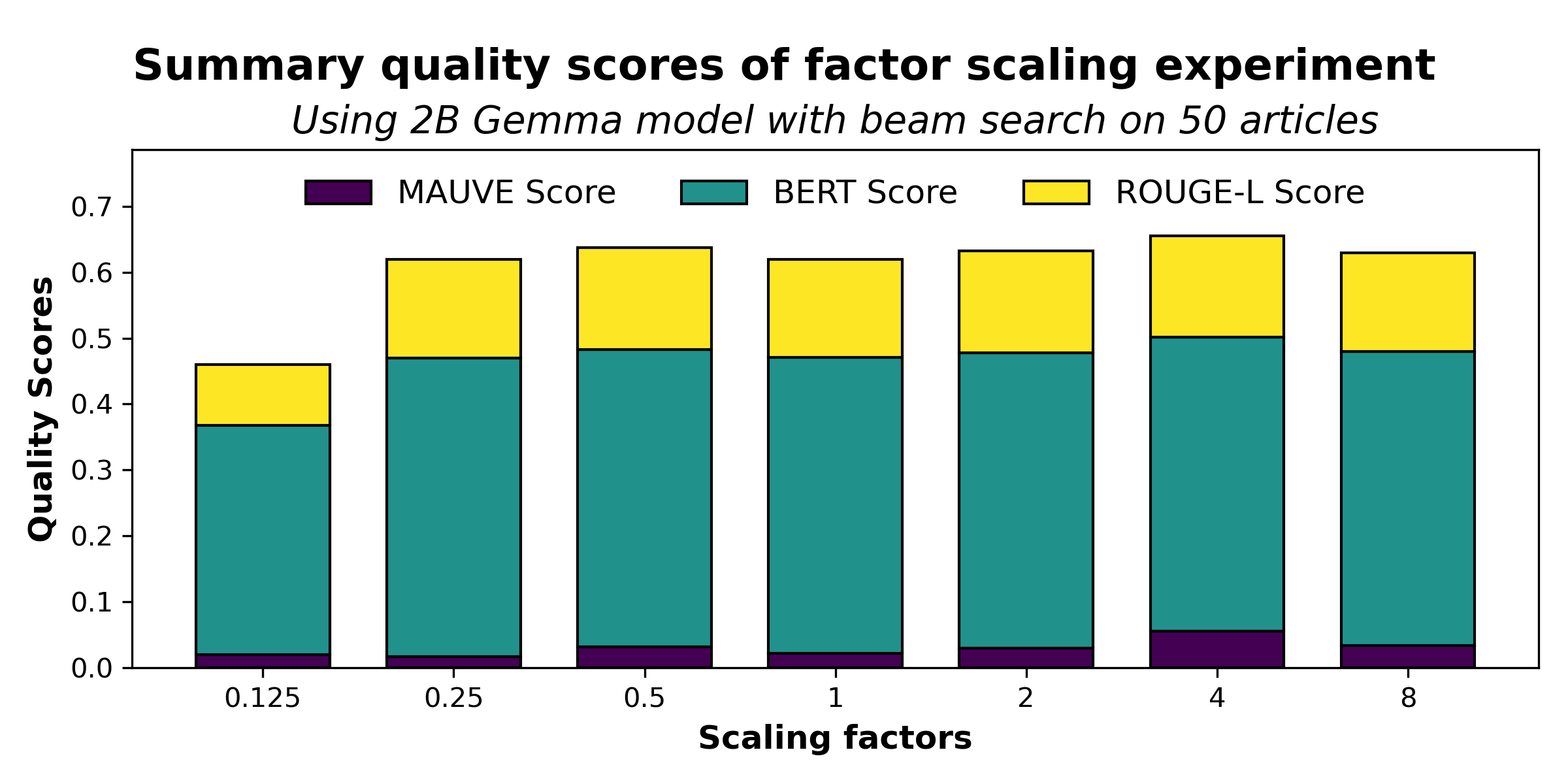}
    \end{minipage}
    \caption{Summary scores for the factor scaling experiment using Gemma-2B with beam search. For beam search, the topical focus is only increased for factors < 1, and the overall control-quality tradeoff is worse than for greedy decoding or the constant shift method.}
    \label{fig:factor_scaling_gemma_2b_beamsearch}
    \vspace{-0.3cm}
\end{figure}
\begin{figure}[H]
    \centering
    \begin{minipage}{0.49\textwidth}
        \includegraphics[width=\linewidth]{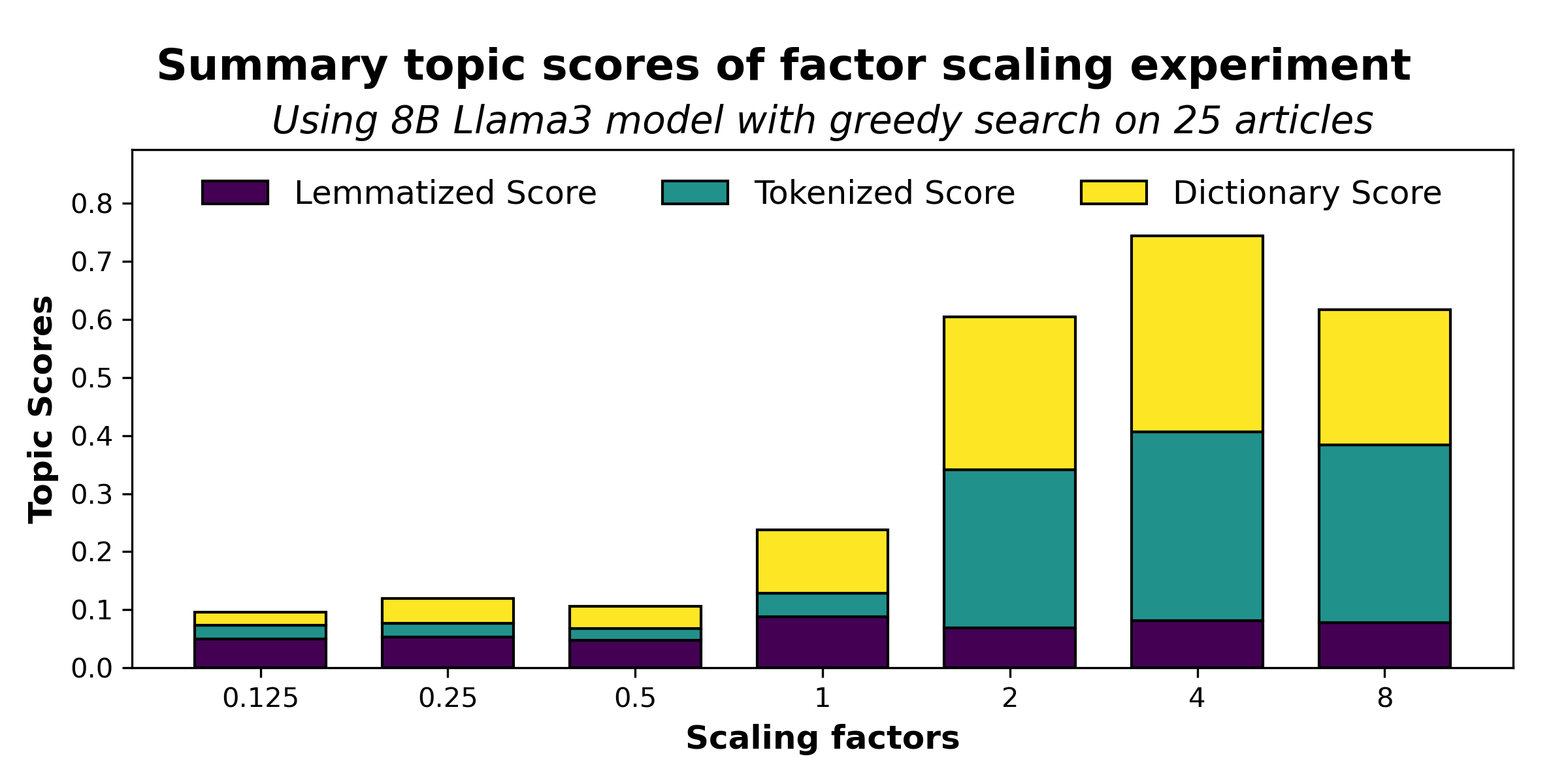}
    \end{minipage}\hfill
    \begin{minipage}{0.49\textwidth}
        \includegraphics[width=\linewidth]{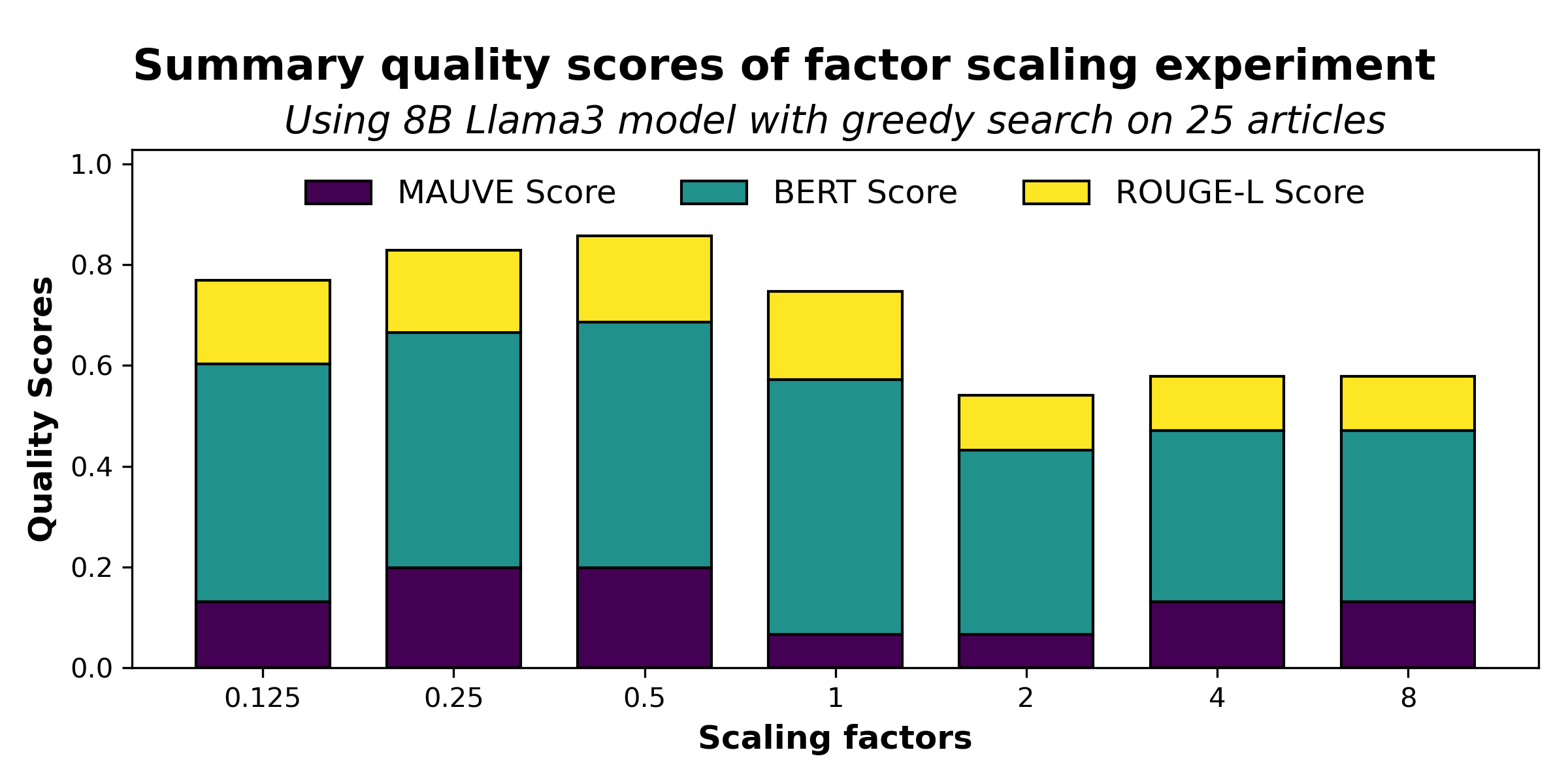}
    \end{minipage}
    \caption{Summary scores for the factor scaling experiment using Llama-3-8B greedy decoding. Because the Llama model outputs positive logits, factor scaling with factors > 1 increases topical focus. The impacts of increased topical focus on summary quality are negative, and the control-quality tradeoff is worse than that of the constant shift method.}
    \label{fig:factor_scaling_llama_8b}
    \vspace{-0.3cm}
\end{figure}
Factor scaling exhibited varying effects across the Llama and Gemma models, influenced by the sign of their logits. For the Gemma model, where logits are predominantly negative, scaling factors below 1 effectively increased the logits, making them less negative and thus more likely to be chosen. Conversely, for positive logits in the Llama model, factors greater than 1 increased logits, enhancing their probability of selection. The use of beam search notably enhanced topical scores only when factors were below 1. Across both models, however, increased topical scores generally correlated with a reduction in summary quality as measured by ROUGE-L, BERTScore, and MAUVE. These results suggest that the impact of scaling factors is contingent on the initial sign of the logits, pointing to the potential benefit of dynamically adjusting factors based on the logit sign for more consistent outcomes across different models.
\subsubsection{Threshold Selection}
\begin{figure}[H]
    \centering
    \begin{minipage}{0.49\textwidth}
        \includegraphics[width=\linewidth]{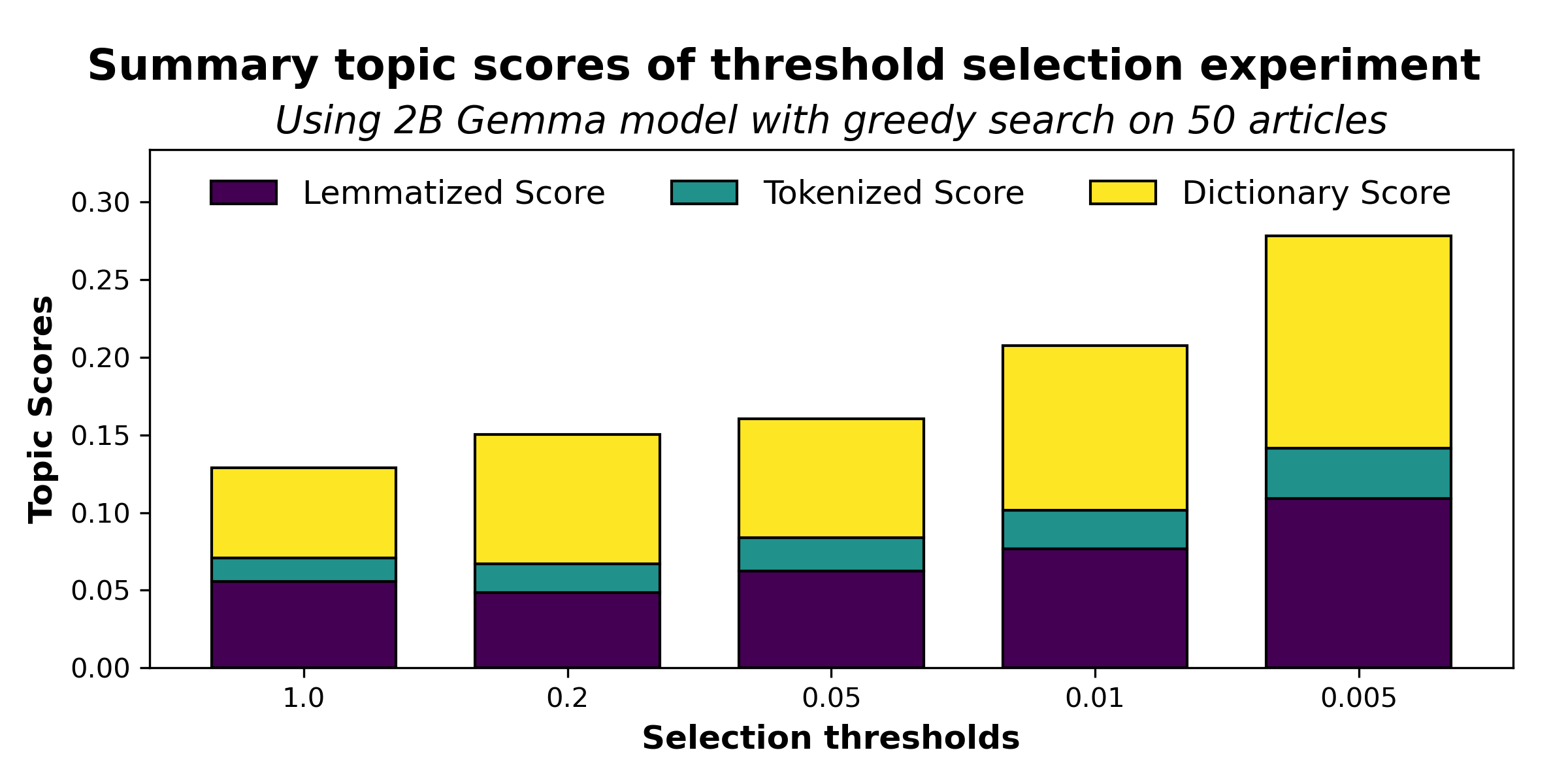}
    \end{minipage}\hfill
    \begin{minipage}{0.49\textwidth}
        \includegraphics[width=\linewidth]{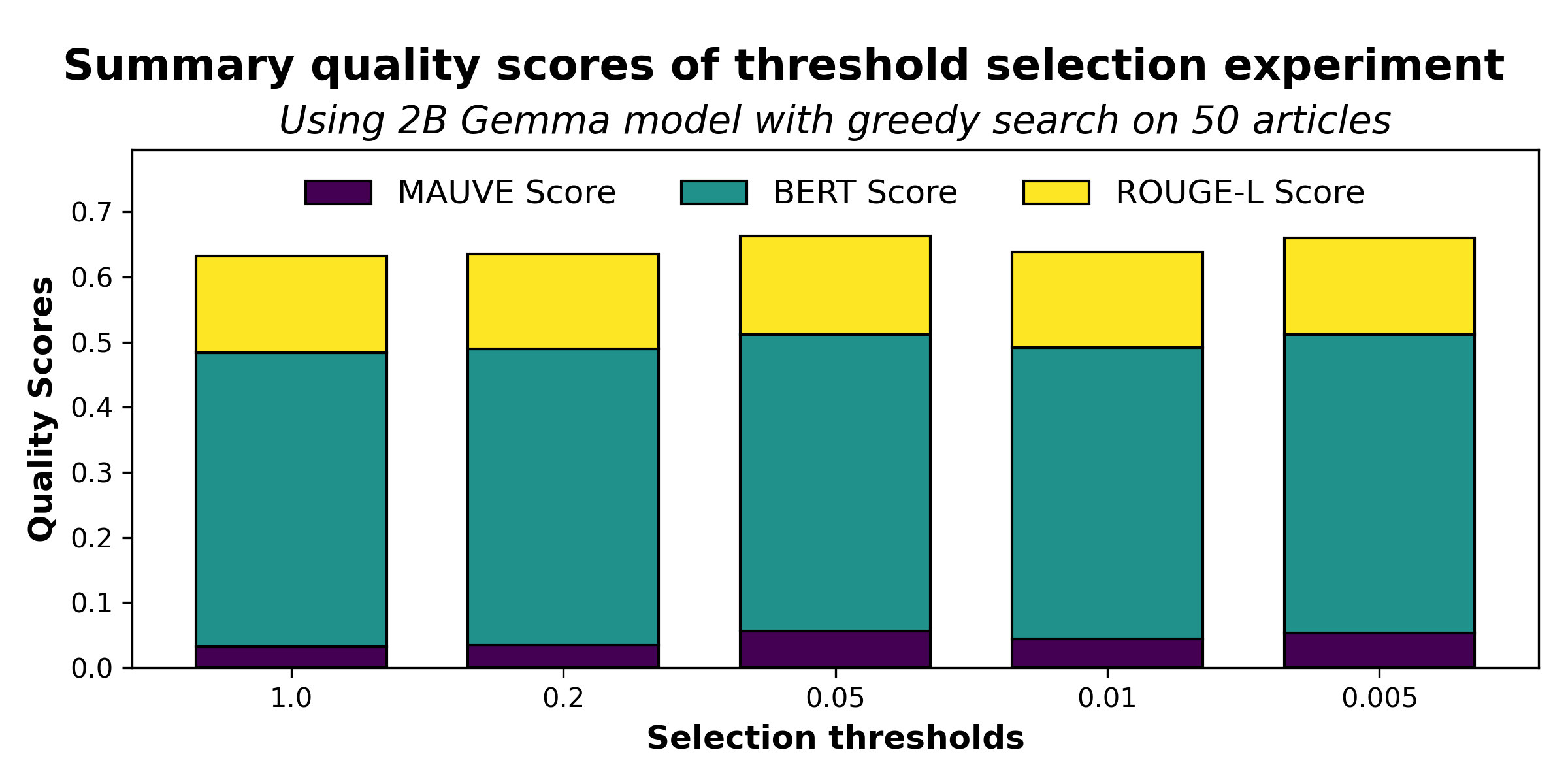}
    \end{minipage}
    \caption{Summary scores for the threshold selection experiment using Gemma-2B with greedy decoding. Topic scores reliably increase with a lower selection threshold without impacting summary quality.}
    \label{fig:threshold_selection_gemma_2b_greedy}
\end{figure}

\begin{figure}[H]
    \centering
    \begin{minipage}{0.49\textwidth}
        \includegraphics[width=\linewidth]{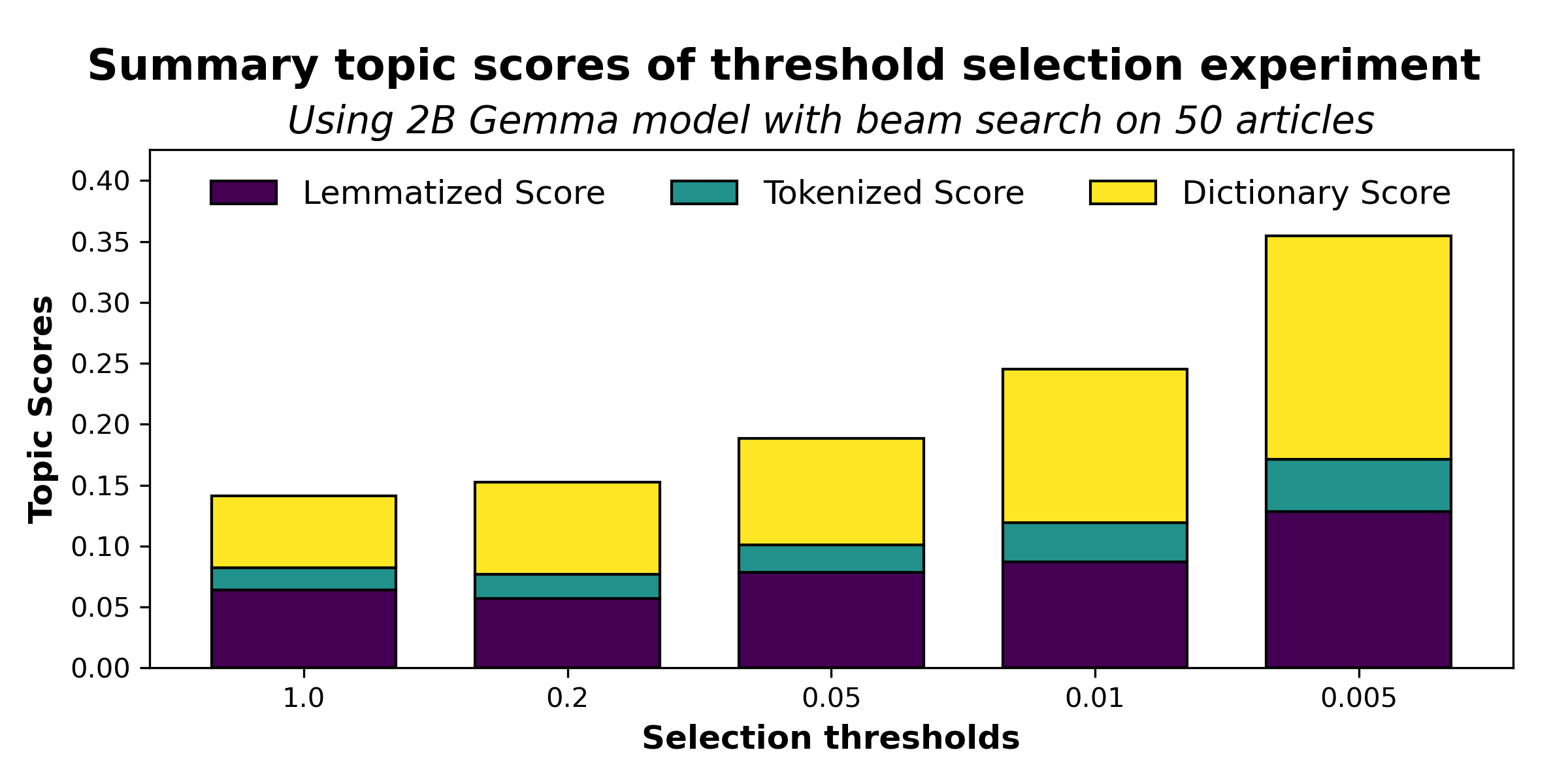}
    \end{minipage}\hfill
    \begin{minipage}{0.49\textwidth}
        \includegraphics[width=\linewidth]{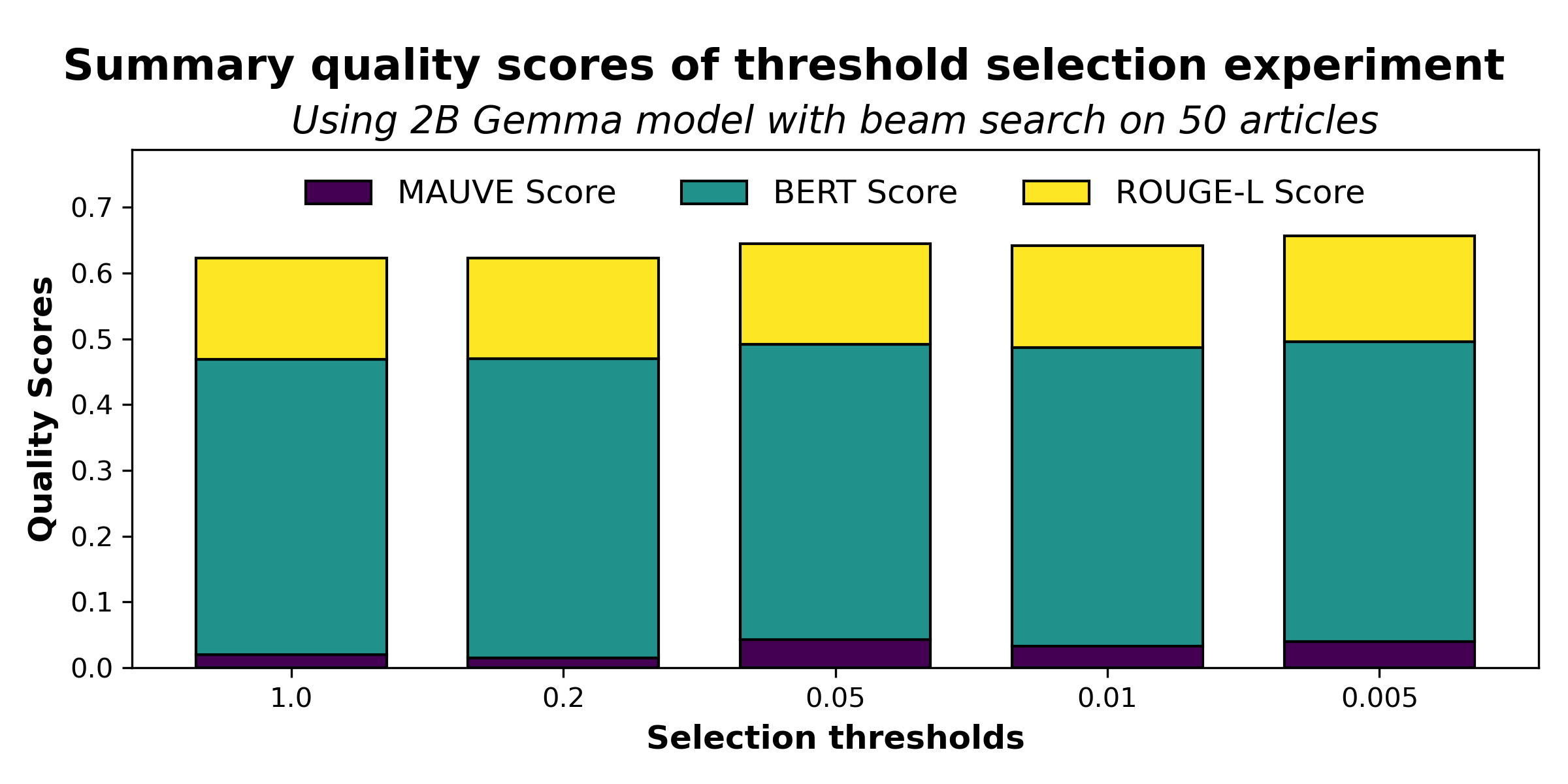}
    \end{minipage}
    \vspace{-0.1cm}
    \caption{Summary scores for the threshold selection experiment using Gemma-2B with beam search. As for greedy search, topic scores increase monotonically for lower selection thresholds, without meaningfully impacting summary quality. Beam search with a selection threshold of 0.005 provides a good control-quality tradeoff for the Gemma-2B model.}
    \label{fig:threshold_selection_gemma_2b_beamsearch}
    \vspace{-0.4cm}
\end{figure}
\begin{figure}[H]
    \centering
    \begin{minipage}{0.49\textwidth}
        \includegraphics[width=\linewidth]{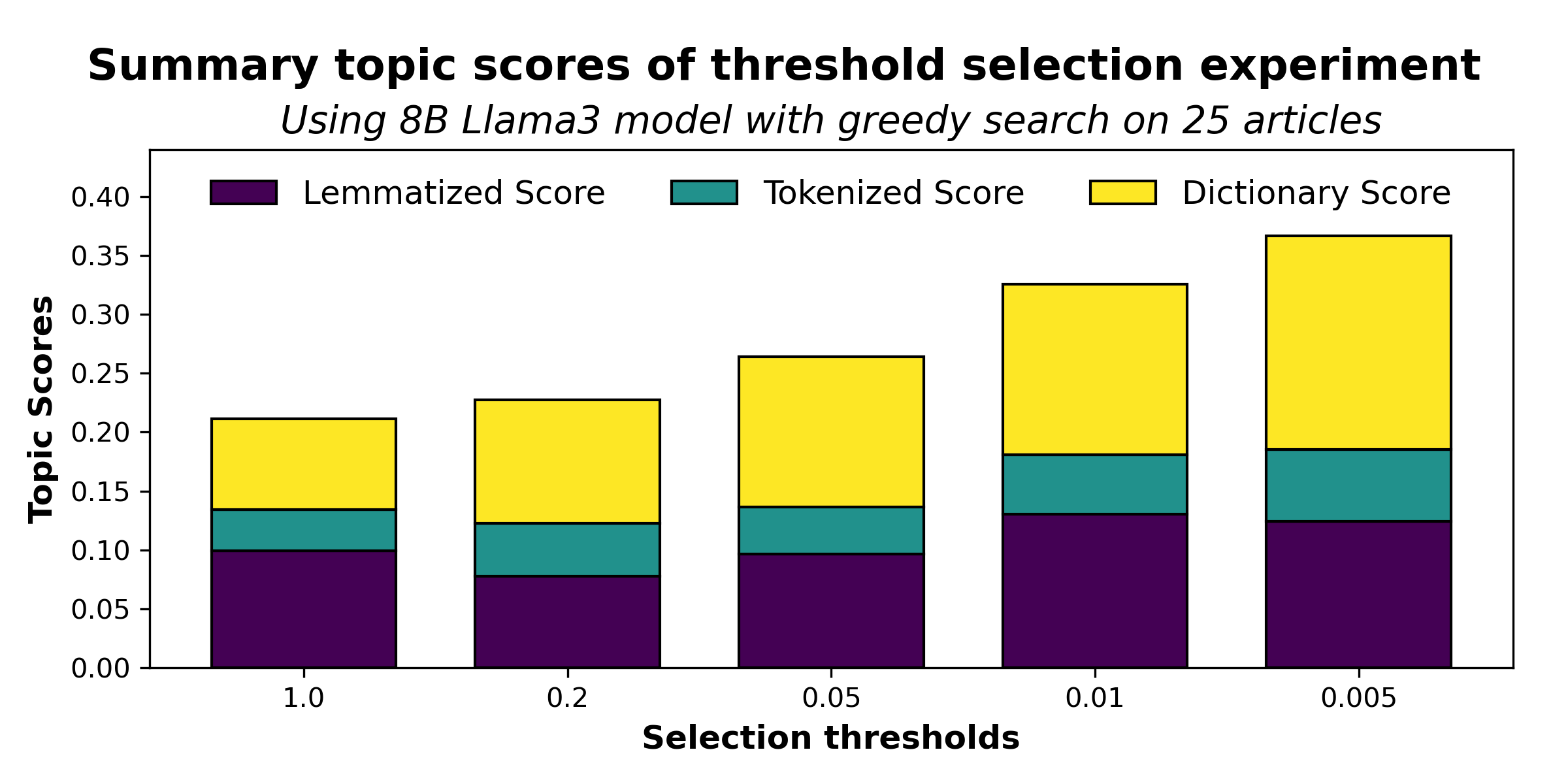}
    \end{minipage}\hfill
    \begin{minipage}{0.49\textwidth}
        \includegraphics[width=\linewidth]{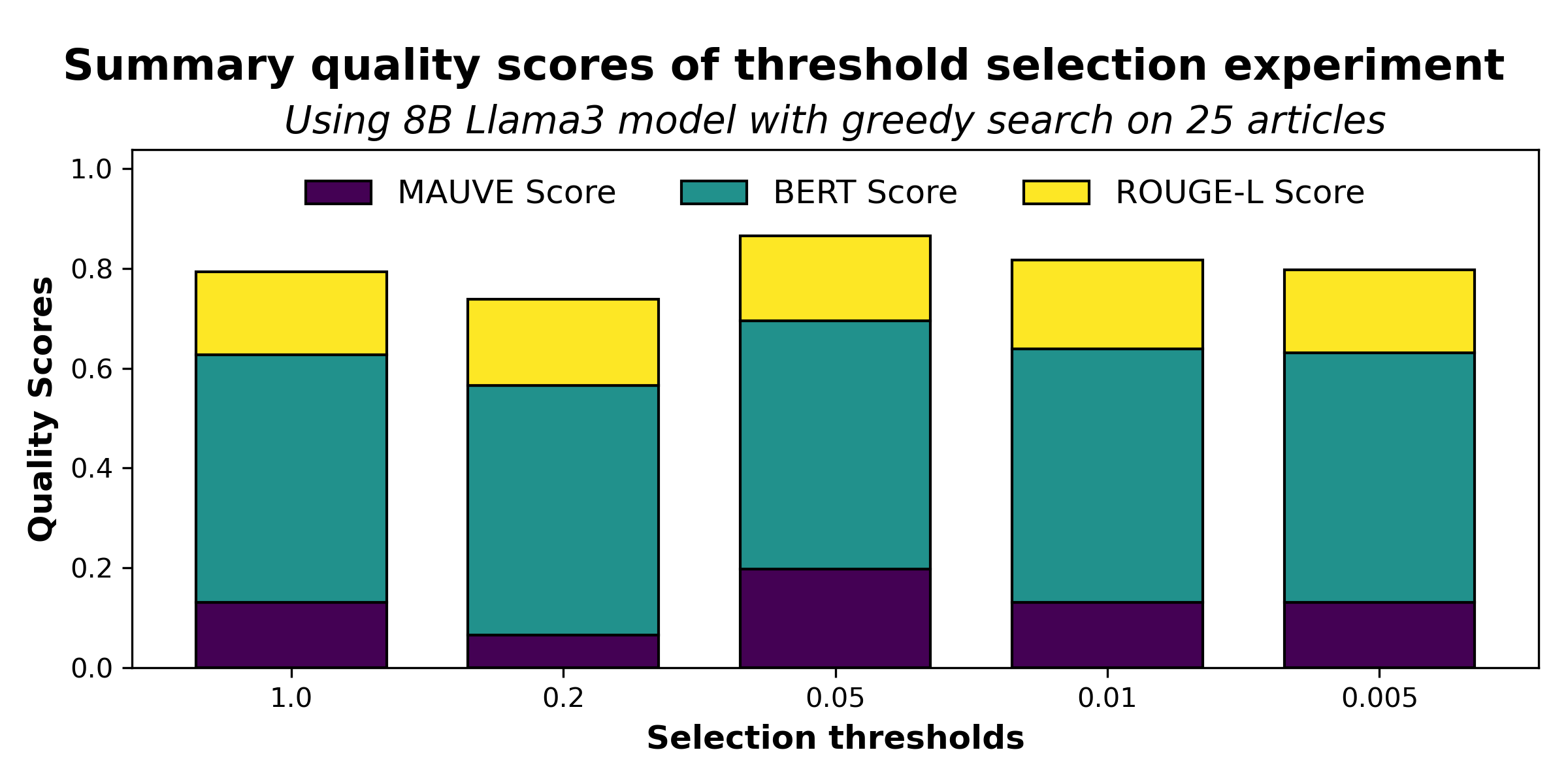}
    \end{minipage}
    \vspace{-0.1cm}
    \caption{Summary scores for the threshold selection experiment using Llama-3-8B with greedy decoding. The Llama model has higher topic and quality scores than the Gemma model. A low selection threshold again provides a good balance between topic control and summary quality.}
    \label{fig:threshold_selection_llama_8b}
    \vspace{-0.3cm}
\end{figure}
Consistent trends were observed in the Threshold Selection experiments across both models and decoding strategies, similar to those seen in the Constant Shift setup. Lowering the probability threshold consistently led to higher topic scores, an effect amplified by beam search because it favors beams with a higher concentration of topic-relevant tokens. The Llama model consistently outperformed the Gemma model in achieving higher summary topic and quality scores. Unlike the Factor Scaling method, increased topical focus through threshold adjustment did not adversely affect summary quality across ROUGE-L, BERTScore, and MAUVE metrics. This indicates that setting topical tokens to the maximum logit value within the current distribution, combined with beam search, yields effective topical control without compromising summary quality.

\section{Discussion}
\subsection{Interpretation of Results}
Our experiments demonstrate that lightweight, inference-time logit interventions can enhance topical focus in abstractive summaries generated by small, instruction-tuned language models. Among the three strategies, Threshold Selection proved to be the most robust: lowering the probability cut‑off steadily increased all three topicality metrics without meaningful degradation in text quality measured by the ROUGE‑L, BERTScore, and MAUVE scores to the reference summaries. Constant Shift also offered smooth controllability, but required model‑specific tuning of the bias magnitude. Factor Scaling was sensitive to the sign of the underlying logits and degraded fluency at a lower control strength, making it the least practical choice. In every setting, the larger Llama‑3‑8B model achieved higher absolute topicality and quality scores than the smaller Gemma-2B, and the 4‑beam search further amplified the gains of logit reweighting. 

Among our tested methods, Threshold Selection emerged as particularly suitable because its probability selection threshold is relatively independent of the logit distribution of the model. This is unlike the Constant Shift and Factor Scaling method, whose hyperparameters require more tuning based on the specific logit distribution of each model. This makes Threshold Selection a preferable choice, as it offers a more universally applicable and less model-dependent approach. Additionally, the use of beam search with Threshold Selection further amplifies its effectiveness, suggesting that this combination may be optimal for improving topical focus in abstractive summaries.

\subsection{Limitations}
While encouraging, these findings come with important caveats: All evaluations were run on the NEWTS dataset~\citep{NEWTS_dataset}, which covers a single news domain, and generalization to other domains and styles remains untested. Results were obtained with one 2B and one 8B parameter instruction-tuned language model, and behavior may differ for larger models, base models, or models with significantly different post-training. Topics were represented by the top 25 LDA words plus inflection variants, so different topic representations may alter outcomes. All quality judgments rely on automatic metrics, and no human evaluation was conducted. Finally, Constant Shift and Factor Scaling require careful tuning because extreme values can compromise fluency.

\subsection{Future Work}
An immediate extension is to adapt the intervention's strength dynamically. Topical steering is unnecessary when the model is about to output function words, such as determiners and conjunctions. By contrast, if the next token is likely to be a content-bearing noun, verb, or adjective, a more substantial bias toward topic-relevant alternatives could improve control. The current threshold method already approximates this behavior by acting only on tokens whose baseline probability exceeds a cutoff, but more sophisticated and still efficient schedules may do even better.

Another promising lightweight approach for controlling text properties during summary generation is the use of steering methods~\citep{Steering_Llama2_via_Contrastive_Activation_Addition} that intervene on intermediate model activations instead of the logits. Empirically, they have been shown to be effective at controlling sentiment~\citep{Self-Refine_Iterative_Refinement_with_Self-Feedback} and other text properties~\citep{A_LMs_guide_through_latent_space}. Their efficacy for free‑form text generation, however, is still largely untested and may be unreliable~\citep{Analyzing_the_Generalization_and_Reliability_of_Steering_Vectors_Daniel_Tan, Sober_look_at_steering_vectors_braun, Understanding_unreliability_of_steering_vectors_in_lms_braun}.

\subsection{Conclusion}
Overall, our findings highlight the potential of logits reweighting as a resource-efficient alternative to more complex model fine-tuning methods, offering a practical solution for enhancing the topical focus of generated abstractive summaries.

\subsection*{Acknowledgements}
This research utilized compute resources at the Tübingen Machine Learning Cloud, DFG FKZ INST 37/1057-1 FUGG.
\bibliography{references}

\begin{thebibliography}{20}
\providecommand{\natexlab}[1]{#1}
\providecommand{\url}[1]{\texttt{#1}}
\expandafter\ifx\csname urlstyle\endcsname\relax
  \providecommand{\doi}[1]{doi: #1}\else
  \providecommand{\doi}{doi: \begingroup \urlstyle{rm}\Url}\fi

\bibitem[Bahrainian et~al.(2021)Bahrainian, Zerveas, Crestani, and Eickhoff]{CATS_Customizable_Abstractive_Topic-based_Summarization}
Seyed~Ali Bahrainian, George Zerveas, Fabio Crestani, and Carsten Eickhoff.
\newblock Cats: Customizable abstractive topic-based summarization.
\newblock \emph{ACM Trans. Inf. Syst.}, 40\penalty0 (1), oct 2021.
\newblock ISSN 1046-8188.
\newblock \doi{10.1145/3464299}.
\newblock URL \url{https://doi.org/10.1145/3464299}.

\bibitem[Bahrainian et~al.(2023)Bahrainian, Jaggi, and Eickhoff]{controllable_topic-focussed_abstractive_summarization}
Seyed~Ali Bahrainian, Martin Jaggi, and Carsten Eickhoff.
\newblock Controllable topic-focused abstractive summarization.
\newblock \emph{CoRR}, abs/2311.06724, 2023.
\newblock URL \url{https://doi.org/10.48550/arXiv.2311.06724}.

\bibitem[Dong et~al.(2023)Dong, Li, Dai, Zheng, Wu, Chang, Sun, Xu, Li, and Sui]{In-context_learning}
Qingxiu Dong, Lei Li, Damai Dai, Ce~Zheng, Zhiyong Wu, Baobao Chang, Xu~Sun, Jingjing Xu, Lei Li, and Zhifang Sui.
\newblock A survey on in-context learning, 2023.

\bibitem[Rafailov et~al.(2023)Rafailov, Sharma, Mitchell, Manning, Ermon, and Finn]{Direct_Preference_Optimization}
Rafael Rafailov, Archit Sharma, Eric Mitchell, Christopher~D Manning, Stefano Ermon, and Chelsea Finn.
\newblock Direct preference optimization: Your language model is secretly a reward model.
\newblock In \emph{Thirty-seventh Conference on Neural Information Processing Systems}, 2023.
\newblock URL \url{https://openreview.net/forum?id=HPuSIXJaa9}.

\bibitem[Ouyang et~al.(2022)Ouyang, Wu, Jiang, Almeida, Wainwright, Mishkin, Zhang, Agarwal, Slama, Gray, Schulman, Hilton, Kelton, Miller, Simens, Askell, Welinder, Christiano, Leike, and Lowe]{RLHF}
Long Ouyang, Jeffrey Wu, Xu~Jiang, Diogo Almeida, Carroll Wainwright, Pamela Mishkin, Chong Zhang, Sandhini Agarwal, Katarina Slama, Alex Gray, John Schulman, Jacob Hilton, Fraser Kelton, Luke Miller, Maddie Simens, Amanda Askell, Peter Welinder, Paul Christiano, Jan Leike, and Ryan Lowe.
\newblock Training language models to follow instructions with human feedback.
\newblock In Alice~H. Oh, Alekh Agarwal, Danielle Belgrave, and Kyunghyun Cho, editors, \emph{Advances in Neural Information Processing Systems}, 2022.
\newblock URL \url{https://openreview.net/forum?id=TG8KACxEON}.

\bibitem[Bahrainian et~al.(2022)Bahrainian, Feucht, and Eickhoff]{NEWTS_dataset}
Seyed~Ali Bahrainian, Sheridan Feucht, and Carsten Eickhoff.
\newblock {NEWTS}: A corpus for news topic-focused summarization.
\newblock In \emph{Findings of the Association for Computational Linguistics: ACL 2022}, pages 493--503, Dublin, Ireland, May 2022. Association for Computational Linguistics.
\newblock \doi{10.18653/v1/2022.findings-acl.42}.
\newblock URL \url{https://aclanthology.org/2022.findings-acl.42}.

\bibitem[Nallapati et~al.(2016)Nallapati, Zhou, dos Santos, Gul{\c{c}}ehre, and Xiang]{CNN_Daily_Mail_dataset}
Ramesh Nallapati, Bowen Zhou, Cicero dos Santos, {\c{C}}a{\u{g}}lar Gul{\c{c}}ehre, and Bing Xiang.
\newblock Abstractive text summarization using sequence-to-sequence {RNN}s and beyond.
\newblock In Stefan Riezler and Yoav Goldberg, editors, \emph{Proceedings of the 20th {SIGNLL} Conference on Computational Natural Language Learning}, pages 280--290, Berlin, Germany, August 2016. Association for Computational Linguistics.
\newblock \doi{10.18653/v1/K16-1028}.
\newblock URL \url{https://aclanthology.org/K16-1028}.

\bibitem[GemmaTeam(2024)]{Gemma_model}
GemmaTeam.
\newblock Gemma: Open models based on gemini research and technology, 2024.

\bibitem[{Llama3Team}(2024)]{Llama3_model}
{Llama3Team}.
\newblock Introducing meta llama 3: The most capable openly available llm to date.
\newblock \url{https://ai.meta.com/blog/meta-llama-3/}, April 2024.
\newblock Accessed: 2024-04-22.

\bibitem[Lin(2004)]{ROUGE_score}
Chin-Yew Lin.
\newblock {ROUGE}: {A} {Package} for {Automatic} {Evaluation} of {Summaries}.
\newblock In \emph{Text {Summarization} {Branches} {Out}}, pages 74--81, Barcelona, Spain, July 2004. Association for Computational Linguistics.
\newblock URL \url{https://aclanthology.org/W04-1013}.

\bibitem[Pillutla et~al.(2021)Pillutla, Swayamdipta, Zellers, Thickstun, Welleck, Choi, and Harchaoui]{MAUVE_score}
Krishna Pillutla, Swabha Swayamdipta, Rowan Zellers, John Thickstun, Sean Welleck, Yejin Choi, and Zaid Harchaoui.
\newblock {MAUVE}: Measuring the gap between neural text and human text using divergence frontiers.
\newblock In A.~Beygelzimer, Y.~Dauphin, P.~Liang, and J.~Wortman Vaughan, editors, \emph{Advances in Neural Information Processing Systems}, 2021.
\newblock URL \url{https://openreview.net/forum?id=Tqx7nJp7PR}.

\bibitem[Zhang* et~al.(2020)Zhang*, Kishore*, Wu*, Weinberger, and Artzi]{BERT_score}
Tianyi Zhang*, Varsha Kishore*, Felix Wu*, Kilian~Q. Weinberger, and Yoav Artzi.
\newblock Bertscore: Evaluating text generation with bert.
\newblock In \emph{International Conference on Learning Representations}, 2020.
\newblock URL \url{https://openreview.net/forum?id=SkeHuCVFDr}.

\bibitem[Devlin et~al.(2019)Devlin, Chang, Lee, and Toutanova]{BERT_model}
Jacob Devlin, Ming-Wei Chang, Kenton Lee, and Kristina Toutanova.
\newblock {BERT}: Pre-training of deep bidirectional transformers for language understanding.
\newblock In \emph{Proceedings of the 2019 Conference of the North {A}merican Chapter of the Association for Computational Linguistics: Human Language Technologies, Volume 1 (Long and Short Papers)}. Association for Computational Linguistics, 2019.
\newblock \doi{10.18653/v1/N19-1423}.
\newblock URL \url{https://aclanthology.org/N19-1423}.

\bibitem[He et~al.(2021)He, Liu, Gao, and Chen]{Deberta_model}
Pengcheng He, Xiaodong Liu, Jianfeng Gao, and Weizhu Chen.
\newblock Deberta: Decoding-enhanced bert with disentangled attention.
\newblock In \emph{International Conference on Learning Representations}, 2021.
\newblock URL \url{https://openreview.net/forum?id=XPZIaotutsD}.

\bibitem[Panickssery et~al.(2024)Panickssery, Gabrieli, Schulz, Tong, Hubinger, and Turner]{Steering_Llama2_via_Contrastive_Activation_Addition}
Nina Panickssery, Nick Gabrieli, Julian Schulz, Meg Tong, Evan Hubinger, and Alexander Turner.
\newblock Steering llama 2 via contrastive activation addition.
\newblock In \emph{Proceedings of the 62nd Annual Meeting of the Association for Computational Linguistics (Volume 1: Long Papers)}. Association for Computational Linguistics, August 2024.
\newblock URL \url{https://aclanthology.org/2024.acl-long.828/}.

\bibitem[Madaan et~al.(2023)Madaan, Tandon, Gupta, Hallinan, Gao, Wiegreffe, Alon, Dziri, Prabhumoye, Yang, Gupta, Majumder, Hermann, Welleck, Yazdanbakhsh, and Clark]{Self-Refine_Iterative_Refinement_with_Self-Feedback}
Aman Madaan, Niket Tandon, Prakhar Gupta, Skyler Hallinan, Luyu Gao, Sarah Wiegreffe, Uri Alon, Nouha Dziri, Shrimai Prabhumoye, Yiming Yang, Shashank Gupta, Bodhisattwa~Prasad Majumder, Katherine Hermann, Sean Welleck, Amir Yazdanbakhsh, and Peter Clark.
\newblock Self-refine: Iterative refinement with self-feedback.
\newblock In \emph{Thirty-seventh Conference on Neural Information Processing Systems}, 2023.
\newblock URL \url{https://openreview.net/forum?id=S37hOerQLB}.

\bibitem[von R\"{u}tte et~al.(2024)von R\"{u}tte, Anagnostidis, Bachmann, and Hofmann]{A_LMs_guide_through_latent_space}
Dimitri von R\"{u}tte, Sotiris Anagnostidis, Gregor Bachmann, and Thomas Hofmann.
\newblock A language model's guide through latent space.
\newblock In \emph{Proceedings of the 41st International Conference on Machine Learning}, ICML'24. JMLR.org, 2024.

\bibitem[Tan et~al.(2024)Tan, Chanin, Lynch, Paige, Kanoulas, Garriga-Alonso, and Kirk]{Analyzing_the_Generalization_and_Reliability_of_Steering_Vectors_Daniel_Tan}
Daniel Chee~Hian Tan, David Chanin, Aengus Lynch, Brooks Paige, Dimitrios Kanoulas, Adri{\`a} Garriga-Alonso, and Robert Kirk.
\newblock Analysing the generalisation and reliability of steering vectors.
\newblock In \emph{The Thirty-eighth Annual Conference on Neural Information Processing Systems}, 2024.
\newblock URL \url{https://openreview.net/forum?id=v8X70gTodR}.

\bibitem[Braun et~al.(2024)Braun, Krasheninnikov, Anwar, Kirk, Tan, and Krueger]{Sober_look_at_steering_vectors_braun}
Joschka Braun, Dmitrii Krasheninnikov, Usman Anwar, Robert Kirk, Daniel Tan, and David~Scott Krueger.
\newblock A sober look at steering vectors for llms.
\newblock In \emph{AI Alignment Forum, November 2024b. URL https://www. alignmentforum. org/posts/QQP4nq7TXg89CJGBh/a-sober-look-at-steering-vectors-for-llms}, 2024.

\bibitem[Braun et~al.(2025)Braun, Eickhoff, Krueger, Bahrainian, and Krasheninnikov]{Understanding_unreliability_of_steering_vectors_in_lms_braun}
Joschka Braun, Carsten Eickhoff, David Krueger, Seyed~Ali Bahrainian, and Dmitrii Krasheninnikov.
\newblock Understanding (un)reliability of steering vectors in language models.
\newblock In \emph{ICLR 2025 Workshop on Foundation Models in the Wild}, 2025.
\newblock URL \url{https://openreview.net/forum?id=qGCp2AYosf}.

\end{thebibliography}
\end{document}